\documentclass[11pt]{article}
\usepackage[utf8]{inputenc}
\usepackage[T1]{fontenc}
\usepackage{lmodern}
\usepackage{amsmath}
\usepackage{amssymb}
\usepackage{graphicx}
\usepackage{booktabs}
\usepackage{array}
\usepackage{longtable}
\usepackage{multirow}
\usepackage[margin=1in]{geometry}
\usepackage{caption}
\usepackage[hidelinks]{hyperref}
\usepackage{xurl}
\usepackage{xcolor}
\usepackage{enumitem}
\usepackage{float}
\usepackage{ragged2e}

\newcommand{\drop}{\texttt{[DROP]}}

\title{Privacy Leakage in Federated Learning in Radiology Reports: A Comparative Evaluation of Tokenizer-Driven Privacy Risks}

\author{
\begin{minipage}{0.9\textwidth}\centering
Santhosh Parampottupadam\textsuperscript{1,2}, Andr\'es Mart\'inez\textsuperscript{1}, Dimitrios Bounias\textsuperscript{1,2}, Sinem Sav\textsuperscript{3}, Klaus Maier-Hein\textsuperscript{1,2,4}, Ralf Floca\textsuperscript{1}
\end{minipage}\\[1.2em]
\normalsize
\begin{minipage}{0.86\textwidth}
\centering
\textsuperscript{1}\,German Cancer Research Center (DKFZ), Division of Medical Image Computing, Heidelberg, Germany\\
\textsuperscript{2}\,Medical Faculty Heidelberg, Heidelberg University, Heidelberg, Germany\\
\textsuperscript{3}\,Department of Computer Engineering, Bilkent University, Universiteler 06800 \c{C}ankaya/Ankara, T\"urkiye\\
\textsuperscript{4}\,Pattern Analysis and Learning Group, Department of Radiation Oncology, Heidelberg University Hospital, 69120 Heidelberg, Germany
\end{minipage}
}

\date{}

\begin{document}
\maketitle
\renewcommand{\thefootnote}{}\footnotetext{Preprint. Under review at \textit{JMIR Medical Informatics}.}\renewcommand{\thefootnote}{\arabic{footnote}}

\section*{Abstract}
\noindent\textbf{Background:} Federated learning (FL) enables multi-institutional model training on clinical text without sharing raw data; however, gradient inversion methods can reconstruct sensitive information from shared model updates. The extent of such privacy leakage in FL applied to radiology reports, and the role of tokenizer design, remains unclear.

\noindent\textbf{Objective:} To quantify gradient-based reconstruction of radiology report text in an FL setting and to compare privacy risk across three transformer tokenization strategies in a controlled, tokenizer-aware evaluation.

\noindent\textbf{Methods:} Six FL clients trained a GPT-2--style transformer (sequence length 32) on two public radiology corpora comprising 368,751 diagnostic reports, 98,206 discharge summaries, and 1,500 MIMIC-CXR free-text reports. Models were trained using three tokenizers (GPT-2, RadBERT, LLaMA-2) with batch sizes of 64, 128, and 256. An active malicious-server threat model was assumed (the server modifies the shared model architecture before distribution), and analytic gradient inversion was applied to recover text. Reconstruction fidelity was measured over five runs using exact sentence accuracy, S-BLEU, G-BLEU, and ROUGE-L.

\noindent\textbf{Results:} Exact sentence reconstruction ranged from 31\% to 44\% across tokenizers (30.6--43.5\% across the 27 tokenizer $\times$ dataset $\times$ batch-size cells). At batch size 64 on the Discharge dataset, accuracy was 42.1\% (GPT-2), 42.3\% (RadBERT), and 39.4\% (LLaMA-2), decreasing to 37.3\%, 37.2\%, and 34.3\% at batch size 256. S-BLEU scores declined with increasing batch size (e.g., GPT-2: 0.44$\rightarrow$0.33; RadBERT: 0.48$\rightarrow$0.35; LLaMA-2: 0.39$\rightarrow$0.30). RadBERT yielded higher reconstruction fidelity and greater recovery of clinical terms, but no tokenizer prevented leakage.

\noindent\textbf{Conclusions:} Substantial portions of radiology report text can be reconstructed from FL gradients even with larger batch sizes and domain-specific tokenizers. Tokenizer design influences leakage severity and should be incorporated into privacy evaluations for clinical language models. Integrating safeguards such as secure aggregation and differential privacy is likely necessary to meet HIPAA and GDPR requirements when deploying FL for radiology NLP.

\noindent\textbf{Trial Registration:} Not applicable.

\noindent\textbf{Keywords:} Federated Learning; Radiology; Privacy; Gradient Inversion; Large Language Models; Data Security; Transformer Models; Patient Confidentiality; HIPAA; GDPR

\section*{Introduction}
Radiology reports serve as an indispensable tool in medical diagnostics, providing critical insights [1] that complement imaging data. These textual data contain rich clinical information, including patient history, diagnostic impressions, and recommended follow-ups, often bridging gaps left by imaging alone. In recent years, advancements in artificial intelligence (AI), particularly transformer-based large language models (LLMs) [2], have revolutionized natural language processing, making it possible to analyze unstructured textual data at scale. LLMs such as the Generative Pretrained Transformer (GPT) family [3] are now being explored for their potential in radiology to assist in generating summaries, extracting relevant clinical insights, and identifying patterns across large datasets [4--6]. A recent systematic review of LLM evaluations in clinical medicine highlights the rapid growth of these models and underscores the need for robust evaluation frameworks to ensure their safety, reliability, and ethical alignment in healthcare applications [7].

Radiology large language models can be collaboratively developed across multiple institutions, leveraging diverse clinical reports to enhance model robustness and improve generalizability across unseen healthcare settings. Nevertheless, healthcare data, such as radiological data sharing, faces significant challenges due to stringent data protection laws like the Health Insurance Portability and Accountability Act (HIPAA) [8] in the U.S. and the General Data Protection Regulation (GDPR) [9] in the EU, which mandate strict safeguards for the privacy of sensitive patient data. Federated Learning (FL) [10], a decentralized machine learning approach where models are trained across multiple institutions without exchanging raw data, has emerged as a promising solution to address these challenges by ensuring that sensitive data remains within each participating institution while still enabling collaborative model development. Multi-Centric FL clinical collaborations [11--14] are transforming medical research by enabling privacy-preserving data sharing, fostering innovation, and addressing critical concerns around data ownership and security.

Nevertheless, FL's decentralized nature introduces inherent vulnerabilities that may compromise patient confidentiality [15,16]. While they eliminate direct data sharing, FL systems rely on the exchange of model parameters between institutions and a central server. An active malicious server --- one that not only observes gradients but also modifies the shared model architecture before distribution --- can exploit these parameters to reconstruct sensitive data, exposing significant vulnerabilities. Existing research has shown that gradient inversion attacks [17--19] can reconstruct private text from models trained on generic language datasets, but such risks remain largely unexplored in the domain of radiology [20].

Recent advances in natural language processing have demonstrated that leveraging domain-specific tokenizers [17] and training LLM on domain-adapted corpora significantly enhances performance in specialized fields such as medicine and genomics. For instance, studies like [21] introduced PubMedBERT, showing that pre-training on biomedical literature yields superior results on downstream clinical tasks compared to generalist models. Similarly, prior studies such as [22] demonstrate that fine-tuning LLMs like BioBERT and MedAlpaca with domain-specific data and tailored tokenization strategies substantially enhances performance in medical tasks. These adaptations enable the models to capture nuanced, domain-relevant semantics that general-purpose LLMs typically overlook.

To specifically investigate the role of tokenization in privacy leakage within federated learning, we isolate tokenizer design as the primary variable while keeping the underlying model architecture constant. While most prior work compares entire language models [7], we argue that vocabulary segmentation alone can influence the susceptibility of models to gradient inversion attacks. Tokenizers trained on domain-specific corpora (e.g., RadBERT) are more likely to encode medical terms as single tokens, increasing their semantic coherence and making them easier to reconstruct. In contrast, general-purpose tokenizers tend to fragment clinical phrases into multiple subwords, which may reduce both interpretability and recoverability. This design choice not only impacts downstream clinical utility [21] but also alters the re-identifiability of sensitive entities during model inversion. By decoupling the tokenizer from the model architecture, our approach provides a targeted evaluation of how vocabulary structure mediates the trade-off between semantic fidelity and patient privacy.

In our study, RadBERT [23], a radiology-specific transformer model, demonstrated consistently higher reconstruction fidelity under attack compared to general-purpose models like GPT-2 [24] and LLaMA-2 [25]. While this highlights the advantages of domain-adapted pretraining for capturing clinical semantics, it also reveals a concerning trade-off: specialized models may be more susceptible to privacy leakage in FL settings. This observation aligns with growing evidence that tokenizer and model design choices can significantly influence both performance and risk in domain-specific applications. Building on this insight, we investigate the extent to which radiology report data can be reconstructed from transformer-based models trained in a simulated FL environment. By orchestrating targeted attacks in a multi-client setup using publicly available radiological datasets, we expose critical vulnerabilities in current FL frameworks. Our study deliberately focuses on the attack surface and does not evaluate defense mechanisms or privacy-preserving strategies. Rather, it aims to characterize worst-case risks and motivate future work toward robust, domain-aware privacy protections for safe deployment of clinical foundation models.

\textit{Scope and Contributions.} This work establishes a foundational analysis of the privacy vulnerabilities inherent in applying transformer-based federated learning (FL) to radiology reports. We focus on characterizing the attack surface by quantifying the reconstruction risk under a worst-case gradient-inversion scenario. To this end, we provide a systematic comparison of information leakage across major tokenizers, namely, GPT-2, RadBERT, and LLaMA-2. Our central contribution is the discovery that domain-specific tokenizers like RadBERT, while improving clinical performance, can paradoxically heighten the risk of leaking sensitive patient data. These findings provide a necessary benchmark and guidepost for developing effective privacy-preserving defenses in medical FL systems. Clinical utility is not directly evaluated in the present study and is inferred from prior work establishing improved downstream performance with domain-specific clinical tokenizers.

\section*{Methods}

\subsection*{Ethical Considerations}
This study is a secondary computational analysis of two publicly available, pre-de-identified clinical text corpora: the Dischargesum dataset [28] and the MIMIC-CXR free-text radiology reports accessed via PhysioNet [29] under the standard PhysioNet credentialed-access Data Use Agreement. No new human-subjects data were collected as part of this work. No identifiable individuals appear in any figure, table, or supplementary material. No participants were recruited and no compensation was provided. Per institutional policy at the German Cancer Research Center (DKFZ) and consistent with PhysioNet's terms of use, secondary computational analysis of these de-identified public corpora qualifies for ethics-board exemption; the original IRB approvals covering the primary collection of MIMIC-CXR (Beth Israel Deaconess Medical Center) and Dischargesum extend to secondary research use without additional consent. All experiments were performed on local DKFZ compute infrastructure and no patient data were transmitted to external services.

\subsection*{Federated Learning and Transformer Models}
Federated Learning (FL) [10] is a decentralized machine learning paradigm that allows multiple institutions to collaboratively train a common model without exchanging raw data. Each client trains locally and sends only gradient updates to a central server for aggregation. In our implementation, we simulated this multi-institutional setup using a GPT-2--style transformer (instantiated as a compact 3-layer encoder with token-embedding dimension $m = 96$, 8 attention heads, feed-forward hidden size 1536, and approximately 13.4\,M trainable parameters, following the Field Guide to Federated Learning baseline architecture [44]) combined with three different tokenizers:

\begin{itemize}[leftmargin=*]
\item GPT-2 tokenizer --- 50,257 tokens
\item StanfordAIMI RadBERT tokenizer --- 30,522 tokens
\item LLaMA-2-7B tokenizer --- 32,000 tokens
\end{itemize}

The overall attack pathway is illustrated in Figure 1.

All tokenizers operated with a fixed sequence length of 32 tokens, so any observed privacy differences are attributable to vocabulary segmentation rather than to input shape or representational scale. Holding the foundation-model architecture fixed at this small, well-characterized baseline allows the tokenizer to serve as the sole independent variable across our GPT-2, RadBERT, and LLaMA-2 comparisons. The 32-token sequence length follows the Field Guide to Federated Learning baseline configuration [44] and is consistent with prior gradient-inversion experiments on text models [41]; it provides a controlled, computationally tractable regime in which the imprint construction's bin disentanglement is well-characterized. To emulate real multi-institutional training, we randomly partitioned two radiology text datasets across six cloud-based clients: the public Dischargesum corpus (98,206 discharge summaries; 368,751 diagnostic reports) and a locally curated sample of 1500 free-text reports from MIMIC-CXR. Training was repeated with batch sizes of 64, 128, and 256 sequences to measure how aggregation granularity affects privacy leakage in shared gradients. Batch sizes refer throughout to 32-token sequences, not whole reports: each report is tokenized and split into multiple non-overlapping 32-token windows; a single discharge summary or radiology report typically contributes between 5 and 40 sequences depending on its length, so the per-client local datasets contain $\ge$ 5,000 sequences and comfortably exceed the largest batch size of 256. Gradient-inversion attacks of the type considered here operate on the gradient produced by a single client during a single training round, which constitutes the standard threat-surface unit in the gradient-leakage literature [42,41]; privacy leakage was therefore quantified per (client, round) rather than over the aggregated federation-level update.

\subsection*{Threat Model}
We characterize the adversary in this study as an active malicious server, following the worst-case threat-model conventions used in prior FL gradient-inversion analyses [43,46]. Formally, the adversary controls the central FL server and possesses the following capabilities. \textbf{(C1) Architecture modification} (active, white-box on the server side): the server may insert auxiliary parameters into the shared model before the first communication round, provided the resulting model still trains end-to-end. We exercise this capability by inserting a single analytic imprint module (Section \textit{Background \& Preliminaries}) immediately before the positional embedding. \textbf{(C2) Per-round per-client gradient observation}: the server observes the full set of weight and bias gradients $\nabla_\theta \mathcal{L}$ returned by each client at each round. We do not assume access to intermediate activations, hidden states, or any non-gradient signal --- i.e., the protocol is standard FedAvg/FedSGD, not split learning. \textbf{(C3) Knowledge of the public model and tokenizer}: the server knows the model architecture (since it distributed it) and the tokenizer's embedding table (since it is part of the shared model). The adversary does not observe raw client data, client-side optimization state, or any cryptographic key material. We assume no defenses are enabled at any layer of the FL stack: no differential privacy, no secure aggregation, no client-side architecture verification, no gradient anomaly detection. This intentionally maximal threat model isolates the upper-bound leakage attributable to gradient inversion alone, which is the quantity our experiments seek to characterize.

\textit{Detectability and stealthiness.} The architecture modification in (C1) is in principle detectable by any client that compares the received model's parameter signature to a published reference. In current production FL frameworks, no such signature verification is mandatory, and small architectural modifications could in principle be disguised by representing the imprint module as a benign feature-extractor or input-normalization layer. However, the imprint module itself adds approximately $(1000 \times 96) + (96 \times 1000) + 1000 \approx 193$\,K parameters --- roughly 1.4\,\% of the 13.4\,M-parameter base model --- and would therefore be detectable by any routine client-side parameter-size audit or cryptographic attestation of the published model signature. We therefore treat detectability as a deployment-policy mitigation, not a fundamental defense, and we recommend in \textit{Practical and Regulatory Implications} that mandatory client-side architecture attestation be considered as part of clinical FL audit protocols.

\textit{Scope of these findings.} The reported reconstruction rates correspond to a worst-case adversary: a malicious central server that (i) owns and modifies the shared model architecture before distributing it to clients, (ii) observes per-round per-client gradients on weight and bias parameters, and (iii) operates in a setting with no privacy defenses enabled --- no differential privacy, no secure aggregation, no gradient clipping, and no client-side model integrity verification. Under typical real-world cross-silo FL deployments, where institutional clients verify the received model architecture, secure aggregation conceals individual client updates, and lightweight differential-privacy noise is applied, these reconstruction rates would be expected to fall substantially. We therefore interpret our numbers as an upper bound on residual leakage when defenses are absent, not as expected leakage in defended deployments. This framing is consistent with the worst-case methodology of prior gradient-inversion studies [42,41].

\subsection*{Background \& Preliminaries}
Let the token-embedding dimension be $m$ and the probe output dimension be $k$. For a single token position, the hidden vector $x \in \mathbb{R}^m$ is produced by the embedding table before positional addition and attention mixing. We insert a small linear probe (an ``imprint'' module in the sense of [42]) on top of this hidden state:

\begin{equation}
y = W_2 \cdot \mathrm{ReLU}(W_0 x + b_0),
\end{equation}

\noindent with $W_0 \in \mathbb{R}^{K \times m}$, $b_0 \in \mathbb{R}^K$, and $W_2 \in \mathbb{R}^{m \times K}$, where $K = k$ denotes the number of probe bins. The probe is appended once at initialization and trained jointly with the rest of the model. As in standard cross-silo federated learning (FedAvg; [10]), each client transmits only the parameter gradient $\nabla_\theta \mathcal{L}$ of its local loss back to the server; no intermediate activations or hidden states are exchanged. This rules out split-learning style leakage and isolates the threat surface to the gradient update itself.

We initialize $b_0$ to the cumulative quantiles of a Laplace prior over the expected pre-activation distribution and $W_0$ to a random Gaussian matrix; $W_2$ is initialized so that $y \approx x$ in expectation, leaving downstream training behaviour essentially unchanged. Under this construction the bins partition the input space into cumulative half-spaces of geometrically decreasing measure, so that for sufficiently large $K$ at most one sample in a mini-batch activates each bin after consecutive-bin differencing. In general, batch aggregation destroys per-sample information: the transmitted gradient $\nabla_\theta \mathcal{L}$ averages over all $B$ samples in the batch and is not invertible without optimization [42]. The cumulative-bin imprint construction circumvents this barrier by analytic design rather than by optimization --- the contribution of each individual token to $\nabla W_0$ and $\nabla b_0$ is structurally separated by bin index $i$, so the server can isolate per-sample gradients even though it only ever observes the batch sum.

From these two captured gradients, the input hidden vector $x$ can be recovered in closed form without optimization:

\begin{equation}
\hat{x}_j = \frac{\nabla W_{0,i}}{\nabla b_{0,i}},
\end{equation}

\noindent where $i$ is the bin uniquely activated by token $j$ after the cumulative-bin differencing step $\nabla W_{0,i} \leftarrow \nabla W_{0,i} - \nabla W_{0,i-1}$ (and analogously for $\nabla b_{0,i}$). We apply this per token and per sample in a batch; sentence reconstructions are formed by decoding each $\hat{x}_j$ to the nearest token in the active tokenizer's embedding table $E \in \mathbb{R}^{V \times m}$ under cosine similarity, where $V$ is the tokenizer vocabulary size. We set $K = 1000$ bins and $m = 96$, with $\hat{x}_j$ L2-normalized before decoding; when $\|\nabla b_{0,i}\|_2$ is below a numerical threshold $\varepsilon$ we add a ridge term $\varepsilon$ to the denominator for numerical stability.

Ties or near-ties (cosine margin $< \delta$) among candidate tokens are broken by re-scoring with the model's language-model-head logits in a server-side forward pass conditioned on the previously-decoded prefix of the same sample --- feasible because the malicious server holds the full shared model and can run forward passes on partial reconstructions. Token $j$ is therefore resolved in left-to-right context rather than in isolation; the very first position falls back to the unconditional logit prior.

\subsection*{Attack Implementation}
We implement the gradient-inversion attack as a deterministic, server-side procedure operating on per-round gradients. The analytic probe and closed-form inversion are defined in Section \textit{Background \& Preliminaries}; below we describe the practical instrumentation, data capture, and decoding steps used in our experiments.

\textit{Step 1 --- Instrumentation.} The server augments the shared transformer with a lightweight linear probe inserted immediately after the token embedding lookup and before positional additions and attention mixing (see \textit{Background \& Preliminaries}). This placement exposes token-level representations prior to any token mixing, ensuring reconstructed vectors reflect tokenizer segmentation rather than later architectural effects.

\textit{Step 2 --- Gradient capture.} During each client training round, the server records the probe gradients returned in the aggregation step --- specifically, the per-round weight gradient $\nabla_{W_0}\mathcal{L} \in \mathbb{R}^{K \times m}$ and bias gradient $\nabla_{b_0}\mathcal{L} \in \mathbb{R}^K$ of the imprint module's first linear layer (notation as in \textit{Background \& Preliminaries}). The ``upstream signal'' $\partial\mathcal{L}/\partial y$ referenced in prior gradient-inversion literature is not transmitted by clients; it is reconstructed server-side from the captured weight and bias gradients under the cumulative-bin construction. These per-round gradients are stored for subsequent inversion and auditing; we retain round identifiers and batch metadata to link reconstructions to training conditions (dataset, tokenizer, batch size).

\textit{Step 3 --- Closed-form inversion and stabilization.} For each captured token gradient in a round, we compute the closed-form embedding estimate $\hat{x}$ following the expression in \textit{Background \& Preliminaries}. After inversion, each $\hat{x}$ is L2-normalized to match embedding norm scaling; when $\|g\|_2$ is near zero we add a small ridge term $\varepsilon$ to the denominator for numerical stability:

\begin{equation}
\hat{x} = \frac{\left(\dfrac{\partial\mathcal{L}}{\partial W}\right)^{\!\top} g}{g^{\top} g + \varepsilon}, \qquad \varepsilon = 10^{-8}.
\end{equation}

\textit{Step 4 --- Token decoding.} Each recovered vector $\hat{x}$ is decoded by nearest-neighbour search over the active tokenizer's embedding table using cosine similarity (top-1). When multiple vocabulary entries are within a small cosine-similarity margin $\delta$ of the maximum, the server breaks the tie by re-scoring candidates using the model's language-model-head logit at that position, computed in a server-side forward pass conditioned on the previously-decoded prefix. This is feasible because the malicious server holds the full shared model (it distributed the model in the first place) and can execute forward passes on partial reconstructions; the very first position falls back to the unconditional logit prior. Decoded tokens are concatenated in positional order to yield sentence-level reconstructions.

\textit{Clarification on positional order recovery.} Positional order is determined by an optimal-assignment matching between recovered breached embeddings and the 32 known positional embeddings of the sequence, implemented as \texttt{scipy.optimize.linear\_sum\_assignment} (the Hungarian algorithm) in the breaching library, following Fowl et al. [42]. The language-model-head logit re-scoring described above operates within each assigned position to resolve vocabulary ambiguity; it does not itself determine position. The phrase ``immediately before the positional embedding'' refers to topological placement within the encoder stack (before the attention and token-mixing layers); the recovered breached embeddings nonetheless carry the positional signature used by the matching stage above.

\textit{Step 5 --- Design for fair tokenizer comparison.} To isolate tokenization effects from architecture or training hyperparameters we hold the model architecture, embedding dimension $m$, probe output dimension $k$, and probe placement constant across experiments. The same inversion and decoding hyperparameters (ridge $\varepsilon$, L2 normalization, nearest-neighbour metric) were applied to GPT-2, RadBERT, and LLaMA-2 tokenizer evaluations without per-tokenizer tuning. Because decoding uses each tokenizer's own embedding table, reconstruction fidelity reflects vocabulary segmentation and tokenizer-induced exposure (single token entities vs fragmented subwords).

\textit{Implementation notes and novelty.} All inversion and decoding steps were implemented in Python (NumPy/PyTorch). Nearest-neighbor lookup used a brute-force cosine search for reproducibility; for large vocabularies, the same pipeline is compatible with approximate nearest-neighbor libraries (FAISS). Per-round artifacts (timestamps, batch composition) were retained to permit round-level analyses and the violin plots reported in Figure 3 and Multimedia Appendices 3--5. The analytic probe itself follows prior work [26]; our novelty lies in (i) its embedding-stack placement to isolate tokenizer-specific effects, (ii) a tokenizer-controlled, architecture-fixed evaluation revealing a privacy--utility trade-off, and (iii) clinical entity-level leakage quantification using MedGemma NER [27], which distinguishes boilerplate text from clinically re-identifiable content.

\textit{Clarification on the cumulative-bin regime.} The $K = 1000$ bins are allocated globally over the batch rather than per token position, so the expected per-bin occupancy $B/K$ is computed over the full pool of $B \times T$ tokens ($T = 32$ here). The strict single-occupancy regime ($B < K$) is exact at our smallest batch ($B = 64$); at the largest configuration ($B = 256$ sequences = 8192 individual tokens) some bins are multiply occupied. The analytic recovery rule $\hat{x}_j = \nabla W_i / \nabla b_i$ is exact only in the single-occupancy regime; under multi-occupancy it yields a per-bin weighted average of the active samples' embeddings, producing a blurred but non-vacuous reconstruction. Positional disambiguation in this regime is performed by the optimal-assignment matching stage described in the \textit{Attack Implementation} (Step 4) operating on the recovered continuous embeddings. The reconstruction-accuracy degradation we report across batch sizes $64 \rightarrow 128 \rightarrow 256$ is consistent with this graceful degradation rather than a strict pigeonhole failure.

\begin{figure}[H]
\centering
\includegraphics[width=\linewidth]{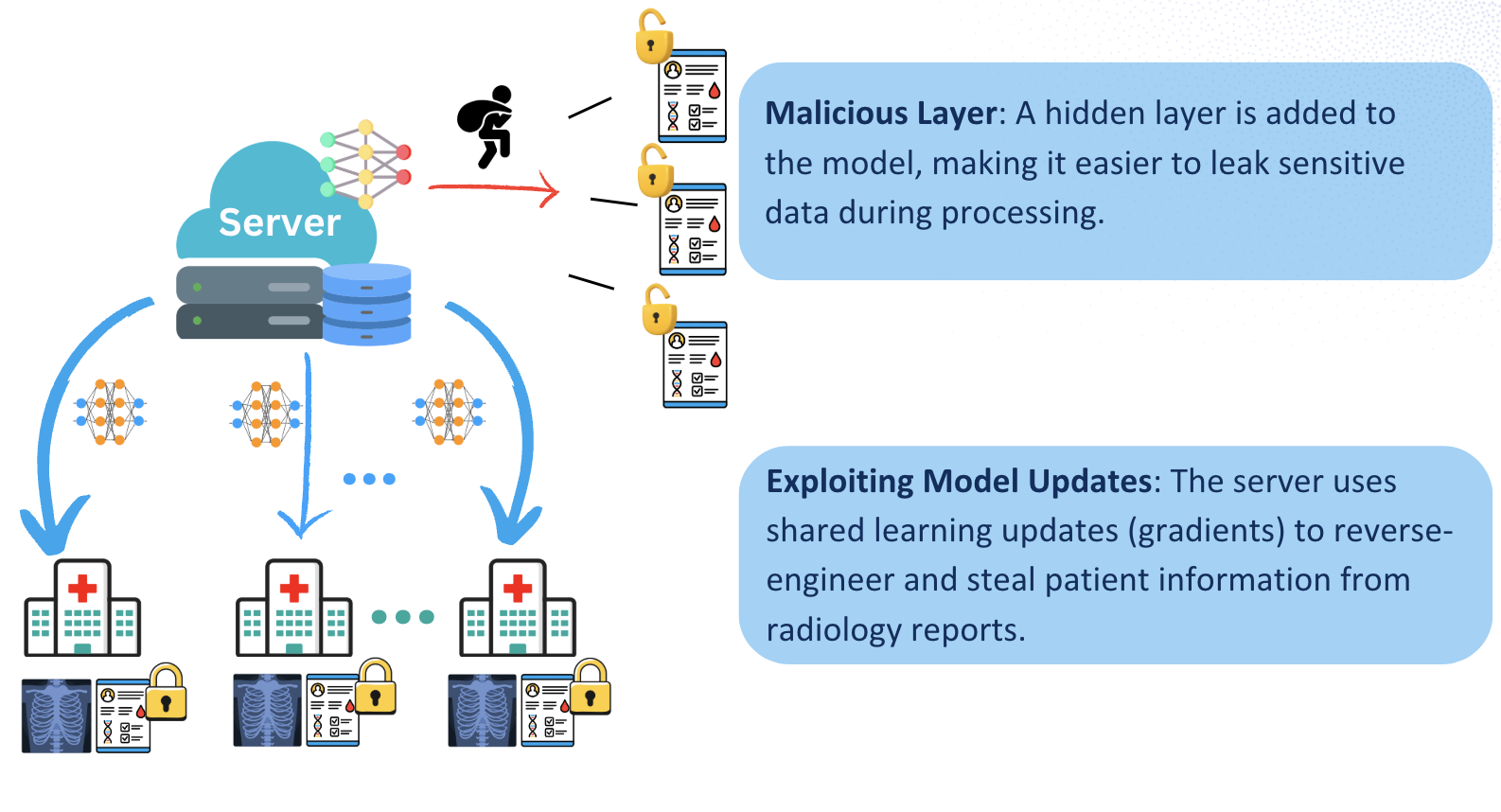}
\caption*{\textbf{Figure 1.} Gradient-inversion attack pathway in federated radiology NLP. A malicious central server coordinates model training across multiple hospitals. By inserting a lightweight linear probe at the embedding stage, the server exposes token-level representations and, from the probe's gradients, reconstructs token embeddings in closed form. Decoding these embeddings via the tokenizer's embedding table yields patient text, illustrating that federated learning can leak confidential report content even without raw data sharing.}
\end{figure}

\subsection*{Dataset and Experimental Setup}
We evaluated privacy leakage using three radiology text corpora. The Dischargesum [28] dataset contained 98,206 discharge summaries and 368,751 diagnostic reports, which were evenly distributed across six simulated institutional clients. To complement these structured reports, we also included a subset of 1,500 free-text MIMIC-CXR [29] radiology reports, similarly partitioned across the six clients.

\textit{MIMIC-CXR sample.} From the full MIMIC-CXR free-text corpus ($\approx$ 227,000 reports), we sampled 1,500 reports uniformly at random without replacement, conditional on each report being non-empty after standard whitespace stripping. No filtering was applied by clinical content, report length, institutional source, or temporal range. The sample identifier list is included in our reproducibility package and available from the corresponding author on reasonable request, subject to the PhysioNet Data Use Agreement.

Each client trained a GPT-2--style transformer locally and transmitted only gradient updates to the central server for aggregation, emulating a standard cross-institutional FL workflow. Reports were uniformly sampled at random across the six clients, producing an approximately IID partition for the purposes of this study. Experiments systematically varied the local batch size (64, 128, and 256 sequences) to evaluate its effect on gradient-level privacy leakage. Every configuration was repeated across five independent runs to account for stochastic training variability and to ensure reproducibility; 95\% confidence intervals across runs are reported in Multimedia Appendix 6; exact paired t-tests on a verification re-run are reported in Multimedia Appendix 8. Training details and reproducibility configuration are reported in Multimedia Appendix 7. At each measurement point, the malicious server captures the gradient transmitted by one client in a given round and applies the analytic reconstruction described in \textit{Background \& Preliminaries} to that single update, in line with the standard single-step formulation adopted in the gradient-inversion literature [42]. Each batch-size configuration was evaluated under this per-client, per-round setting across the five independent runs.

\subsection*{Evaluation Metrics}
We assessed reconstruction fidelity using four complementary metrics chosen to capture both exact and partial text recovery, following established practice in prior gradient-inversion and text reconstruction studies [13--16, 23]. These metrics together quantify the extent to which patient-identifiable information can be re-created from shared model updates.

\begin{itemize}[leftmargin=*]
\item \textbf{Exact Sentence Accuracy.} The percentage of reconstructed sentences that match the ground truth exactly, reflecting the proportion of complete exposures of patient text. A reconstructed sentence is counted as an exact match if, after the following normalization steps, every token at every position equals the corresponding token in the original sentence: (i) Unicode NFKC normalization; (ii) lower-casing; (iii) collapsing all runs of internal whitespace to a single space; (iv) preserving punctuation as part of the token sequence rather than stripping it. Sentences are extracted from the concatenated stream of decoded 32-token windows by segmenting at terminal punctuation; the metric is therefore evaluated on these reconstructed sentence units rather than directly on the raw 32-token chunks. We report a stricter case-sensitive variant on the same five-run experiment for completeness; the qualitative ordering of tokenizers is unchanged across both variants. Exact-sentence accuracy is reported as the conservative upper bound on full-sentence exposure: token-level partial recovery (a single recovered diagnostic noun phrase, for instance) can constitute clinically meaningful leakage even when the full sentence is not reconstructed exactly; this regime is captured by the $n$-gram and longest-common-subsequence metrics described below. The two metric families together bracket the true leakage rate.

\item \textbf{Sentence-Level BLEU (S-BLEU)} [30--31]: Calculates $n$-gram precision, where an $n$-gram is a contiguous sequence of $n$ words (e.g., unigrams, bigrams, trigrams, and four-grams). BLEU compares 1--4 word sequences in the reconstructed sentence against the reference, with a brevity penalty to discourage overly short outputs. Higher scores indicate stronger local overlap and preservation of clinical phrasing.

\item \textbf{ROUGE-L} [33]: Measures recall based on the longest common subsequence (LCS) between reconstruction and reference, capturing the extent to which key clinical terms and their order are retained in longer spans.

\item \textbf{Global BLEU (G-BLEU)} [30]: Aggregates all reconstructed and reference texts into a single corpus and computes BLEU precision across the entire set. BLEU precision measures how many $n$-gram sequences (1--4 words) in the reconstructed text are also present in the reference text, reflecting the accuracy of content reproduction. By evaluating over the full corpus rather than sentence-by-sentence, G-BLEU captures overall reconstruction consistency and helps identify systematic leakage patterns in model outputs.
\end{itemize}

By combining these four metrics, we quantify both the frequency of perfect reconstructions and the degree of partial text recovery, offering a comprehensive assessment of patient data exposure under our federated learning attack.

\textit{Choice of metrics.} BLEU and ROUGE-L were originally proposed for translation and summarization quality, where their numerical values are bounded by human-level $n$-gram overlap. In the gradient-inversion context, we use them as information-recovery metrics rather than as quality metrics: a higher BLEU or ROUGE-L between a reconstructed sentence and the original indicates more shared $n$-gram content --- i.e., a greater amount of original wording recovered. Exact-sentence accuracy provides the strict-recovery upper bound; S-BLEU and ROUGE-L provide the partial-recovery signal that captures cases where most of a sentence is recovered but a small number of tokens are mis-decoded. This complementary use of strict and partial metrics is consistent with prior gradient-inversion text studies [45,41]. We deliberately do not adopt semantic-similarity metrics such as BERTScore in the main results because they can credit paraphrastic matches that do not constitute literal information leakage, which is the privacy question of interest here.

\subsection*{Named Entity Recall Evaluation}
We quantified recovery of clinically meaningful content using the Google MedGemma medical NER model [27]. MedGemma was run on a random sample of 100 original discharge summaries to build a reference vocabulary of 1,440 unique entity surface forms (diagnoses, procedures, anatomical structures, medications). The same model was applied to reconstructed summaries for each tokenizer configuration. The 100-report sample size was chosen on the basis of clinical-entity vocabulary saturation: in pilot internal analyses the cumulative count of unique MedGemma-extracted entities plateaus near 1,400 unique surface forms by sample = 70 reports and is approximately flat by sample = 100, indicating that this sample is sufficient to characterize the recoverable clinical-entity vocabulary at the corpus level. Larger samples (e.g., 500+ reports) would yield diminishing returns in unique-entity coverage at substantially higher computational cost.

\textit{Normalization and matching.} Entity strings were lowercased and whitespace-trimmed; internal punctuation was preserved. An entity was counted as recovered if its surface form exactly matched any term in the 1,440-item reference (case-insensitive). No fuzzy matching, synonym expansion, or manual adjudication was used.

\textit{Metric.} For each tokenizer, we report Recovered Terms (unique count) and Recall Rate (\%): Recall Rate (\%) = (Recovered Terms / 1,440) $\times$ 100. We also report the union across tokenizers (``Total Unique Terms Recovered'') computed against the same 1,440-term reference.

\section*{Results}

\subsection*{Quantitative Reconstruction Success}
Exact sentence-level reconstruction accuracy on the Discharge dataset was:

\begin{itemize}[leftmargin=*]
\item GPT-2 tokenizer: 42\% / 40\% / 37\% at batch sizes 64 / 128 / 256
\item RadBERT tokenizer: 42.3\% / 40\% / 37\%
\item LLaMA-2 tokenizer: 39.4\% / 37\% / 34\%
\end{itemize}

On MIMIC-CXR, reconstruction ranged 33\%--41\% sentences (Figure 2), showing that neither tokenizer choice nor larger batch sizes fully mitigates data leakage. Detailed accuracy results are provided in Multimedia Appendix 1. All reported reconstruction metrics represent the mean of five independent runs; 95\% confidence intervals across runs are reported in Multimedia Appendix 6; exact paired t-tests on a verification re-run are reported in Multimedia Appendix 8. Across all configurations, standard deviations were below 1 percentage point, confirming that the observed trends are statistically stable.

\subsection*{Impact of Batch Size on S-BLEU}
Increasing the batch size led to a clear reduction in average reconstruction fidelity across all metrics (Table 2; Figure 3). For discharge reports, mean S-BLEU decreased from 0.44 at a batch size of 64 to 0.33 at 256; diagnosis reports fell from 0.43 to 0.32; and MIMIC-CXR from 0.45 to 0.32. Parallel trends were observed for ROUGE-L and G-BLEU, confirming that coarser gradient aggregation generally mitigates privacy leakage. Notably, across all datasets and batch sizes, the RadBERT tokenizer consistently achieved higher reconstruction scores than GPT-2 and LLaMA-2. This suggests that domain-specific tokenization may facilitate more accurate recovery of clinical language---enhancing semantic fidelity but potentially increasing privacy risk. Corresponding mean S-BLEU, G-BLEU, and ROUGE-L scores are summarized in Multimedia Appendix 2.

\begin{figure}[H]
\centering
\includegraphics[width=0.82\linewidth]{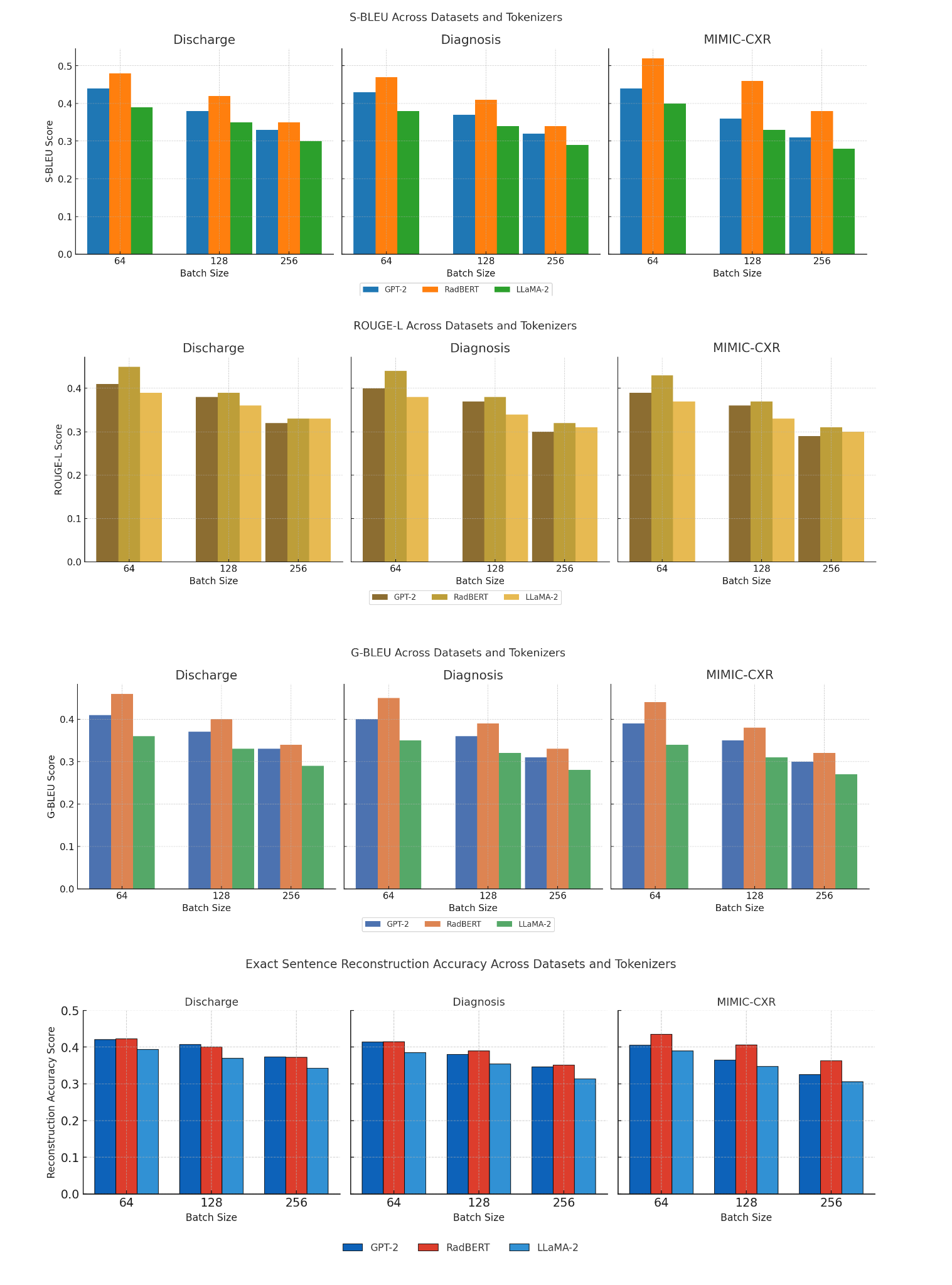}
\caption*{\textbf{Figure 2.} Reconstruction fidelity across datasets and tokenizers at varying batch sizes. Top row: S-BLEU scores; second row: ROUGE-L scores; third row: G-BLEU scores; bottom row: Exact Sentence Reconstruction Accuracy scores. Each column corresponds to a dataset (Discharge, Diagnosis, and MIMIC-CXR), and colors indicate tokenizers (GPT-2, RadBERT, and LLaMA-2). Across all metrics and datasets, RadBERT consistently achieves higher reconstruction fidelity than GPT-2 and LLaMA-2, with LLaMA-2 showing the weakest performance. Increasing the batch size from 64 to 256 reduces reconstruction quality across all models, illustrating a trade-off between training granularity and patient data privacy.}
\end{figure}

\subsection*{Variability in Gradient Inversion Severity Across Training Rounds}
While average reconstruction fidelity declined with larger batch sizes across all metrics, individual training rounds revealed substantial fluctuations in leakage severity. We illustrate this variability using S-BLEU in Figure 3, as it provides a balanced measure of local $n$-gram precision and sentence-level structure, making it particularly sensitive to round-to-round differences. However, similar variability patterns were also observed for Exact Sentence Reconstruction Accuracy, ROUGE-L, and G-BLEU, which are summarized in Multimedia Appendices 3--5. Even at the highest batch size (256), certain rounds produced high-fidelity reconstructions comparable to those at batch size 64, underscoring a persistent privacy risk. Coarser gradient aggregation therefore reduces but does not eliminate patient information exposure in federated learning.

In real-world medical AI deployments, even a single outlier round with high leakage could compromise patient confidentiality. Therefore, it is essential to characterize not just average-case leakage but also variability across rounds, to ensure that system-level guarantees account for worst-case scenarios.

\begin{figure}[H]
\centering
\includegraphics[width=0.72\linewidth]{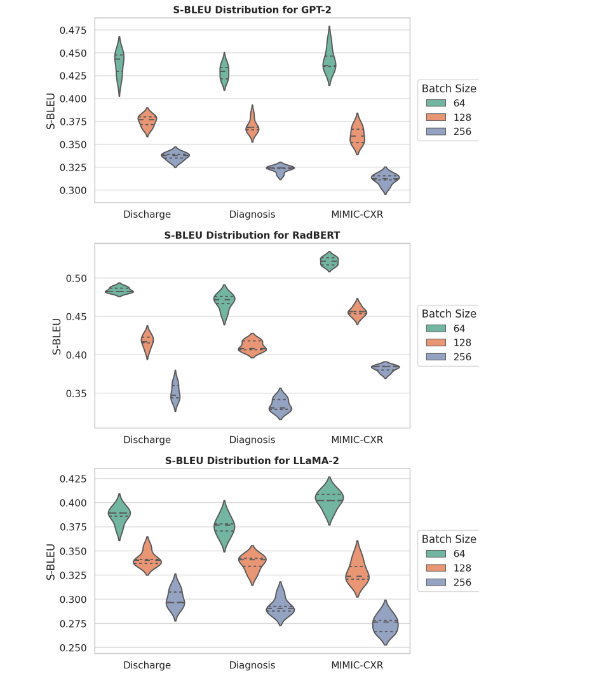}
\caption*{\textbf{Figure 3.} Violin plots showing the distribution of S-BLEU scores across five runs for each tokenizer (GPT-2, RadBERT, and LLaMA-2) on three datasets: Discharge, Diagnosis, and MIMIC-CXR. Each violin represents performance at a given batch size (64, 128, 256), with color-coded distinctions shown in the legend. The plots illustrate both central tendency and variability, highlighting that RadBERT consistently achieved higher reconstruction fidelity, while LLaMA-2 exhibited greater spread and lower median performance. Distributions compress as batch size increases, reflecting reduced sentence-level recoverability.}
\end{figure}

\subsection*{Clinical Term Recall in Reconstructed Text}
As presented in Table 4, RadBERT reconstructions recovered the highest number of clinical terms (260, or 18.1\%), followed by GPT-2 (180 terms, 12.5\%) and LLaMA-2 (136 terms, 9.4\%). In total, 380 unique terms were leaked across all models, corresponding to a cumulative recovery rate of 26.4\%.

These findings demonstrate that gradient inversion attacks can reveal not only structural sentence fragments but also clinically meaningful entities, posing a tangible privacy risk. The elevated recall from RadBERT, despite its clinical utility, indicates that domain adaptation may heighten vulnerability in federated training settings.

\subsection*{Qualitative Reconstruction}
Tables 1--3 showcase original and reconstructed outputs from the Dischargesum, Radiology, and MIMIC-CXR datasets, highlighting tokenizer-specific vulnerabilities. GPT-2 reconstructions retained general sentence structure but frequently dropped key clinical terms. RadBERT, by contrast, preserved more medically relevant entities such as ``catheter'' and ``nodularity,'' reflecting its domain-specific tokenization. LLaMA-2 exhibited the most fragmented outputs, with high token dropout and less consistent semantic coherence. These qualitative results reinforce the quantitative findings, demonstrating that even domain-adapted transformer models remain susceptible to gradient-based inversion attacks, recovering not only template phrases but also clinically meaningful patient information.

\begin{table}[H]
\small
\centering
\caption*{\textbf{Table 1.} Discharge Report: Reconstruction across tokenizers (GPT-2, LLaMA-2, and StanfordAIMI/RadBERT). The original report and reconstructions are aligned by model. Recovered content varies significantly: RadBERT and GPT-2 preserve more clinical semantics, while LLaMA-2 drops more tokens, indicating weaker reconstruction performance. \textit{Note on alignment.} The \drop{} marker indicates token positions for which the gradient-inverted reconstruction did not produce a valid token (a bin collision or numerical-instability bin). For visual clarity these positions are aligned against the original text using oracle alignment, which the attacker does not have access to in a real attack setting; the attacker would observe an unordered or partially ordered set of recovered tokens without knowledge of where reconstruction failed. Oracle alignment is used here purely for reader interpretability and does not reflect adversary capability.}
\begin{tabular}{@{}>{\RaggedRight}p{2.1cm}>{\RaggedRight\arraybackslash}p{12.3cm}@{}}
\toprule
\textbf{Model} & \textbf{Text} \\
\midrule
Original Report & Dear Mr.\ Alex, It Was A Pleasure Taking Care Of You Here At ABC clinic. You Were Admitted To Our Hospital After Undergoing Repair Of Your Ventral Hernia. You Have Recovered From Surgery And Are Now Ready To Be Discharged To Home With Services. Please Follow The Recommendations Below To Ensure A Speedy And Uneventful Recovery. ACTIVITY: - Do not drive until you have stopped taking pain medicine and feel you could respond in an emergency. - You may climb stairs. - You may go outside, but avoid traveling long distances until you see your surgeon at your next visit. \\
\addlinespace
GPT-2 & Dear Mr.\ Alex, It Was A \drop{} Taking Care Of You \drop{} At ABC clinic. You Were Admitted To Our Hospital After Undergoing Repair Of Your Ventral Hernia. You Have Recovered From Surgery And Are Now Ready To Be Discharged To Home With Services. Please Follow The Recommendations \drop{} To Ensure A And \drop{} Recovery. ACTIVITY: - Do \drop{} drive until you have \drop{} taking pain medicine and feel you could respond in an emergency. - You may climb stairs. - You may go outside, but avoid traveling \drop{} distances until you \drop{} your surgeon at your next visit. \\
\addlinespace
LLaMA-2 & \drop{} Mr.\ Alex, It \drop{} A Pleasure Taking Care Of You Here At ABC clinic. You Were Admitted To Our Hospital After Undergoing Repair Of Your Ventral \drop{}. You Have \drop{} From \drop{} And \drop{} \drop{} \drop{} To Be Discharged To Home With Services. Please Follow The \drop{} Below To Ensure A Speedy And Uneventful Recovery. ACTIVITY: - Do not drive until you have stopped taking pain \drop{} and feel you \drop{} \drop{} in an emergency. - \drop{} \drop{} climb stairs. - You may go outside, but avoid traveling long distances until you see your surgeon at your next visit. \\
\addlinespace
StanfordAIMI/ RadBERT & Dear Mr.\ Alex, It Was A \drop{} Taking Care Of You Here At ABC clinic. You Were Admitted To Our Hospital After Undergoing Repair Of Your Ventral Hernia. You Have Recovered From \drop{} And Are Now Ready To Be Discharged To Home With Services. Please Follow The Recommendations Below To Ensure A Speedy And Uneventful Recovery. ACTIVITY: - Do not drive until you have stopped taking pain medicine and feel you could \drop{} in an \drop{}. - You may climb stairs. - You may go outside, but avoid traveling \drop{} \drop{} until you see your \drop{} at your next visit. \\
\bottomrule
\end{tabular}
\end{table}

\begin{table}[H]
\small
\centering
\caption*{\textbf{Table 2.} Radiology Report: Comparison of reconstructed output across tokenizers. Despite dense clinical structure, RadBERT and GPT-2 maintain more contextual phrasing than LLaMA-2. Sensitive findings such as ``cirrhosis'' and ``nodularity'' were partially reconstructed in all models. \textit{Note on alignment.} The \drop{} marker indicates token positions for which the gradient-inverted reconstruction did not produce a valid token (a bin collision or numerical-instability bin); four-asterisk sequences (****) in the Original Report row are preserved as native source de-identification redactions from the discharge/radiology reports. For visual clarity these positions are aligned against the original text using oracle alignment, which the attacker does not have access to in a real attack setting; the attacker would observe an unordered or partially ordered set of recovered tokens without knowledge of where reconstruction failed. Oracle alignment is used here purely for reader interpretability and does not reflect adversary capability.}
\begin{tabular}{@{}>{\RaggedRight}p{2.1cm}>{\RaggedRight\arraybackslash}p{12.3cm}@{}}
\toprule
\textbf{Model} & \textbf{Text} \\
\midrule
Original Report & EXAMINATION: LIVER OR GALLBLADDER US (SINGLE ORGAN) INDICATION: History: with cirrhosis, increased abdominal pain TECHNIQUE: Grey scale and color Doppler ultrasound images of the right upper quadrant were obtained. COMPARISON: Abdominal ultrasound from **** FINDINGS: The liver is extremely coarse and nodular in echotexture similar to the prior examination consistent with a history of cirrhosis. Parenchymal heterogeneity limits detection of focal lesions. \\
\addlinespace
GPT-2 & \drop{}: LIVER OR GALLBLADDER US (SINGLE ORGAN) INDICATION: History: with \drop{}, increased abdominal pain TECHNIQUE: Grey scale and color \drop{} ultrasound images of the right upper quadrant were obtained. COMPARISON: Abdominal ultrasound from **** FINDINGS: The liver is \drop{} coarse and \drop{} in echotexture similar to the prior examination consistent with a history of \drop{}. Parenchymal heterogeneity limits detection of focal \drop{}. \\
\addlinespace
LLaMA-2 & EXAMINATION: \drop{} OR GALLBLADDER US (\drop{} ORGAN) \drop{}: History: with cirrhosis, \drop{} abdominal pain TECHNIQUE: Grey scale and color \drop{} ultrasound images of the right upper quadrant were obtained. COMPARISON: Abdominal ultrasound from **** FINDINGS: The liver is extremely \drop{} and nodular in \drop{} similar to \drop{} prior examination consistent with a history of cirrhosis. Parenchymal \drop{} limits detection of focal lesions. \\
\addlinespace
StanfordAIMI/ RadBERT & EXAMINATION: LIVER OR GALLBLADDER US (SINGLE ORGAN) INDICATION: History: with cirrhosis, increased abdominal pain TECHNIQUE: Grey scale and color \drop{} ultrasound images of the right upper quadrant were obtained. \drop{}: Abdominal ultrasound from **** FINDINGS: The liver is extremely \drop{} and nodular in echotexture similar to the prior examination consistent with a history of cirrhosis. \drop{} heterogeneity limits detection of focal lesions. \\
\bottomrule
\end{tabular}
\end{table}

\begin{table}[H]
\small
\centering
\caption*{\textbf{Table 3.} MIMIC-CXR Report: Reconstruction outputs across tokenizers. GPT-2 and RadBERT captured anatomical and pathological keywords like ``fibrosis'' and ``catheter'' more faithfully than LLaMA-2. Structural fidelity was higher in RadBERT outputs, suggesting tokenizer vocabulary influences reconstruction clarity. \textit{Note on alignment.} The \drop{} marker indicates token positions for which the gradient-inverted reconstruction did not produce a valid token (a bin collision or numerical-instability bin). For visual clarity these positions are aligned against the original text using oracle alignment, which the attacker does not have access to in a real attack setting; the attacker would observe an unordered or partially ordered set of recovered tokens without knowledge of where reconstruction failed. Oracle alignment is used here purely for reader interpretability and does not reflect adversary capability.}
\begin{tabular}{@{}>{\RaggedRight}p{2.1cm}>{\RaggedRight\arraybackslash}p{12.3cm}@{}}
\toprule
\textbf{Model} & \textbf{Text} \\
\midrule
Original Report & 10439781,55811525,``Frontal and lateral views of the chest were obtained. Left-sided Port-A-Catheter is similar in position, terminating at the cavoatrial/right atrial junction. Patient has diffuse increase in interstitial markings bilaterally consistent with patient's underlying history of chronic interstitial lung disease with likely overlying pulmonary edema improved since \_\_\_, but similar in appearance as compared to \_\_\_. No definite focal consolidation or pleural effusion. Multilevel vertebroplasties are seen along the thoracic spine, similar to prior.'', Pulmonary edema superimposed on known lung fibrosis. \\
\addlinespace
GPT-2 & 10439781,55811525,``Frontal and \drop{} views of the chest were obtained. Left-sided Port-A-\drop{} is similar in position, terminating at the cavoatrial/right atrial junction. Patient has diffuse increase in interstitial markings bilaterally consistent with patient's underlying history of chronic \drop{} lung disease \drop{} likely overlying pulmonary edema improved since \_\_\_, but similar in \drop{} as compared to \_\_\_. No definite focal consolidation or \drop{} effusion. Multilevel vertebroplasties are seen along the thoracic spine, \drop{} to prior.'', Pulmonary edema superimposed on known lung fibrosis. \\
\addlinespace
LLaMA-2 & 10439781,\drop{},``Frontal and lateral \drop{} of the chest were \drop{}. Left-sided Port-A-Catheter is similar in position, terminating at the cavoatrial/right atrial junction. Patient has diffuse increase in interstitial markings \drop{} consistent with patient's underlying history of chronic interstitial lung \drop{} with likely \drop{} pulmonary edema improved since \_\_\_, but similar in \drop{} as compared to \_\_\_. No \drop{} focal consolidation or pleural \drop{}. Multilevel vertebroplasties are seen along the thoracic spine, similar to prior.'', Pulmonary edema superimposed on known lung \drop{}. \\
\addlinespace
StanfordAIMI/ RadBERT & 10439781,55811525,``Frontal and \drop{} views of the chest were obtained. Left-sided Port-A-Catheter is similar in position, terminating at the cavoatrial/right atrial junction. Patient has diffuse increase in interstitial markings bilaterally consistent with patient's underlying history of chronic interstitial lung disease with likely overlying pulmonary edema improved since \_\_\_, but similar in appearance as compared to \_\_\_. No \drop{} focal consolidation or pleural effusion. \drop{} vertebroplasties are seen along the thoracic spine, similar to prior.'', Pulmonary edema \drop{} on known lung fibrosis. \\
\bottomrule
\end{tabular}
\end{table}

\begin{table}[H]
\centering
\caption*{\textbf{Table 4.} Recovery of clinically meaningful terms from gradient-inverted reconstructions using GPT-2, LLaMA-2, and RadBERT tokenizers. Clinical named entities were identified via the Google MedGemma model across 100 reconstructed discharge summaries and compared against a reference set of 1,440 unique terms.}
\begin{tabular}{@{}lrr@{}}
\toprule
\textbf{Model} & \textbf{Recovered Terms} & \textbf{Recall Rate (\%)} \\
\midrule
GPT-2 & 180 & 12.5\% \\
LLaMA-2 & 136 & 9.4\% \\
RadBERT & 260 & 18.1\% \\
\midrule
Total (Unique) & 380 & 26.4\% \\
\bottomrule
\end{tabular}
\end{table}

\section*{Discussion}

\subsection*{Principal Results}
This study demonstrates that even with domain-specific tokenization and larger batch sizes, transformer models in an FL setup leak substantial patient information. The active server reconstructed up to 42\,\% of radiology report sentences: GPT-2 yielded 42\% / 40\% / 37\% accuracy at batch sizes 64 / 128 / 256; RadBERT, 42.3\% / 40\% / 37\%; and LLaMA-2, 39.4\% / 37\% / 34\%. Sentence-level S-BLEU scores similarly declined (e.g., GPT-2 from 0.44 $\rightarrow$ 0.38 $\rightarrow$ 0.33), but even at the largest batch sizes, nearly one-third of sentences were fully recovered. Across all batch sizes and datasets, RadBERT consistently achieved higher reconstruction scores than GPT-2 and LLaMA-2---for example, reaching an S-BLEU of 0.52---indicating improved fidelity in recovering sentence structure and phrasing. RadBERT preserved more clinical keywords (e.g., anatomical and diagnostic terms) than the generalist models, yet did not eliminate leakage. These findings confirm that neither increasing batch sizes nor specialized tokenizers alone can ensure patient privacy in FL-trained LLMs.

To further disambiguate templated language from truly clinical leakage, we applied MedGemma NER to 100 reconstructed summaries. Across all models, 380 clinical terms were recovered out of 1440 unique entities. These findings confirm that privacy risks extend beyond boilerplate phrasing to include clinical concepts relevant to diagnosis and care.

Our findings underscore that tokenizer design plays a critical role in privacy leakage under gradient inversion attacks. While RadBERT's domain-specific tokenizer improved the retention of clinically meaningful terms, it also increased the recoverability of sensitive information, leaking 260 out of 1440 reference entities (18.1\%) compared to 12.5\% and 9.4\% for GPT-2 and LLaMA-2, respectively. This highlights a fundamental trade-off: the same vocabulary segmentation that enhances clinical utility may inadvertently make protected health information (PHI) easier to reconstruct. In this light, tokenizer choice should be treated not merely as a performance optimization, but as a privacy-sensitive design decision. Future privacy audits and risk assessments of federated clinical models should explicitly account for tokenization strategies, especially as domain-specific LLMs become more prevalent in medical AI.

\textit{Mechanism --- why domain-specific tokenization heightens reconstruction.} The reconstruction step decodes each recovered token-embedding vector back to the nearest token in the active tokenizer's embedding table. A clinical concept such as ``pneumothorax'' is encoded as a single token in RadBERT's domain-specific vocabulary, so a successful nearest-neighbour decode yields the entire concept verbatim. The same concept is segmented into multiple subwords (e.g., ``p / neu / moth / orax'') in GPT-2's byte-pair encoding; recovering the concept therefore requires all of its subword embeddings to be reconstructed correctly and placed in the correct positional order. Each fragmented subword introduces an additional point of failure, and any one mis-decode yields a corrupted concept. Domain-specific tokenization thus reduces the attack-surface diameter per clinical entity --- a property that improves training utility on clinical text but, by the same mechanism, raises gradient-inversion vulnerability.

\textit{Clinical concepts versus directly identifiable PHI.} The MedGemma named-entity recognition model used in this study captures clinical concepts (diagnoses, procedures, anatomical structures, medications) rather than direct identifiers under the HIPAA Safe Harbor enumeration (names, MRNs, dates more granular than year, full-face photographic images, etc.). Direct identifiers are largely already redacted in the public Dischargesum and MIMIC-CXR corpora before release. The recovered terms therefore reflect re-identification potential through clinical context --- for example, an unusually rare diagnosis, a unique procedure pattern, or a specific medication regimen that, in combination with auxiliary recovered context, could plausibly contribute to re-identifying a patient --- rather than direct exposure of an enumerated identifier. We make this distinction explicit because the regulatory framing of ``PHI'' is broader than ``direct identifier''; HIPAA's expert-determination standard (45 CFR \S164.514(b)(1)) treats clinical context recoverable from a record as part of identifiability risk, and the same logic is reflected in the GDPR's notion of indirect identifiability via singling out.

\textit{Why batch size reduces leakage in this attack.} The imprint module's analytic recovery rule $\hat{x}_j = \nabla W_{0,i} / \nabla b_{0,i}$ relies on the assumption that bin $i$ is activated by at most one sample in the batch. With $K = 1000$ bins and batch size $B$, the expected number of samples per bin is $B/K$, and the probability that any given bin is hit by $\ge 2$ samples grows approximately quadratically in $B$ (a birthday-paradox argument). When two or more samples share a bin, the recovered ratio $\nabla W_{0,i} / \nabla b_{0,i}$ becomes a gradient-weighted average of distinct embeddings rather than a clean single-sample recovery, and the corresponding nearest-neighbour decode is corrupted. This explains why exact-sentence accuracy decays with batch size while never reaching zero: as long as some bins remain singly occupied, the corresponding samples are recovered cleanly. The same argument predicts that increasing $K$ relative to $B$ would partially compensate for batch growth --- a direction we leave to future work.

\textit{Key Contribution.} We provide the first systematic evaluation of how different transformer tokenizers impact gradient-inversion vulnerability on radiology text, showing that domain-specific tokenization (RadBERT) may heighten leakage risk even as it better preserves clinical terminology.

\subsection*{Comparison With Prior Work}
Our results extend prior demonstrations of gradient-inversion attacks in generic natural language processing models [17,18] and imaging domains [19] to the structured and template-rich text of radiology reports. Earlier works such as DLG [47] and iDLG [26] reconstructed generic sentences or pixel-level images from shared gradients but did not examine how tokenizer design mediates privacy leakage. Subsequent privacy-preserving FL frameworks in medical imaging [17--19] primarily focused on architectural defenses rather than language-specific vulnerabilities. Unlike these approaches, our study isolates the tokenizer as a controllable privacy variable by holding the transformer architecture and training hyperparameters constant across GPT-2, RadBERT, and LLaMA-2. This tokenizer-controlled setup reveals that vocabulary segmentation alone can significantly influence gradient-based reconstruction fidelity. This comparison against prior gradient-inversion literature is summarized in Table 5. While Akinci et al [20] first discussed text leakage risks in clinical LLMs, no prior work systematically compared multiple tokenization schemes or evaluated reconstruction across both discharge and imaging-report corpora. Our findings therefore provide the first quantitative evidence that domain-adapted tokenizers---though beneficial for clinical utility---may paradoxically amplify privacy leakage in federated radiology LLMs.

\textit{Practical significance.} Earlier closed-form gradient inversion attacks on text models report exact-sentence recovery rates of approximately 30--50\,\% at small batch sizes on generic English corpora [45,41]; our radiology-domain experiments fall within this range (31--44\,\%), confirming that the threat extends from generic to clinical NLP without amplification at the strict-recovery level. The practical significance is, however, qualitatively different in the clinical setting: radiology corpora encode clinical context (diagnoses, procedure types, anatomical findings) at higher density per token than newswire or web text, so equivalent reconstruction rates translate to higher clinical informativeness per recovered sentence. A single recovered sentence in a clinical context may contain a rare diagnosis or unusual procedure combination that contributes meaningfully to re-identification risk, whereas a single recovered sentence from generic web text typically carries no analogous re-identifying signal.

\textit{Scope of the tokenizer comparison.} Because we hold the foundation-model architecture fixed at a small, well-characterized baseline, our findings reflect the effect of vocabulary segmentation on gradient-inversion leakage, not the privacy properties of the GPT-2, RadBERT, or LLaMA-2 foundation models themselves. Larger frontier-scale foundation models exhibit different gradient sparsity and noise dynamics that can change inversion vulnerability in either direction; we do not claim that our results extend to those regimes, and explicit characterization at frontier scale is an important target for future work.

\begin{table}[H]
\footnotesize
\centering
\caption*{\textbf{Table 5.} Comparison with prior gradient-inversion studies. To our knowledge, this is the first study to explicitly hold the foundation-model architecture constant and vary only the tokenizer, in the clinical radiology NLP domain.}
\begin{tabular}{@{}>{\RaggedRight}p{2.2cm}>{\RaggedRight}p{2.3cm}>{\RaggedRight}p{2.4cm}>{\RaggedRight}p{2.1cm}>{\RaggedRight}p{2.1cm}>{\RaggedRight\arraybackslash}p{2.4cm}@{}}
\toprule
\textbf{Study} & \textbf{Model} & \textbf{Tokenizer controlled?} & \textbf{Domain} & \textbf{Datasets} & \textbf{Defense evaluation?} \\
\midrule
{[}47{]} (DLG) & LSTM, ResNet & No (single tokenizer per task) & Generic vision + NLP & CIFAR, MNIST & None \\
\addlinespace
{[}26{]} (iDLG) & LeNet & No & Generic vision & CIFAR & None \\
\addlinespace
{[}46{]} (TAG) & BERT & Single tokenizer & Generic NLP & CoLA, SST-2 & None \\
\addlinespace
{[}42{]} (Robbing the Fed) & Transformer / ViT & No & Vision + WikiText & WikiText, ImageNet & DP discussed \\
\addlinespace
{[}19{]} & UNet, ViT & N/A (vision) & Medical imaging & BraTS, LIDC & Discussed \\
\addlinespace
{[}20{]} & Generic LLMs & Discussed, not measured & Clinical NLP & Review article & --- \\
\addlinespace
\textbf{This work} & 3-layer Transformer (held constant) & \textbf{Yes} --- three tokenizers compared & Clinical radiology NLP & Dischargesum, MIMIC-CXR & Discussed; full evaluation noted as future work \\
\bottomrule
\end{tabular}
\end{table}

\subsection*{Practical and Regulatory Implications}
\textit{Focus on Attack Characterization.} This study deliberately isolates the gradient-inversion attack vector and does not evaluate or compare privacy defenses. Instead, we quantify the worst-case leakage risk under an unconstrained adversary. Subsequent work should build on these findings to empirically evaluate and optimize defense mechanisms in realistic radiology FL pipelines. Our results demonstrate that generic federated learning defenses such as differential privacy [34], secure aggregation [35] and gradient-anomaly detection have yet to be empirically validated for the structured, high-risk text found in radiology reports. Injecting noise via differential privacy may reduce reconstruction fidelity but requires careful calibration to avoid degrading critical diagnostic language. Similarly, cryptographic secure aggregation can obscure individual updates but may introduce prohibitive latency in hospital networks. Real-time gradient-anomaly detection promises early warning of inversion attacks, yet its thresholds must be tuned to the unique update patterns of clinical text to prevent both false alarms and missed breaches. To ensure compliance with HIPAA and GDPR, healthcare organizations may benefit from running pilot evaluations of these defenses under realistic conditions and document performance trade-offs. Regulatory bodies could consider developing formal audit protocols for federated learning systems, encompassing threat modeling, reconstruction testing, and defense efficacy, before granting approval for clinical deployment.

\subsection*{Defenses and Practical Mitigations}
While this study focused on characterizing gradient inversion risk, effective deployment of federated radiology models requires integrating strong privacy defenses [35]. Common mitigation strategies such as synthetic data generation and differential privacy (DP) alone remain insufficient for large language models [36--38]. Synthetic data can leak statistical artifacts and fail to preserve downstream clinical performance [36,37], whereas DP often requires large privacy budgets to maintain utility, limiting its protective value [39].

In practice, more robust protection arises from hybrid approaches that combine secure aggregation to mask individual updates with task-tuned DP noise, gradient clipping, and auditable privacy monitoring. Cryptographic protocols such as homomorphic encryption or secure multiparty computation offer mathematically proven safeguards but may introduce computational overhead in hospital networks [40]. Future implementations of federated clinical NLP should adopt layered defenses that balance efficiency, regulatory compliance, and measurable privacy guarantees [35,40].

\section*{Limitations}
This study was designed as a worst-case evaluation, assuming a fully active malicious-server capable of modifying model architecture and exploiting gradient updates. While extreme, this assumption is appropriate for international or cross-consortium federated settings, where governance may be inconsistent and insider threats cannot be ruled out.

Several constraints limit the generalizability of our findings:

\begin{itemize}[leftmargin=*]
\item \textbf{Dataset Scope.} We used publicly available radiology corpora (Dischargesum and MIMIC-CXR), which lack multimodal components (e.g., images), free-text notes, and operational metadata such as timestamps or identifiers. Future evaluations should test gradient inversion on richer, production-like datasets to better estimate clinical leakage risk.

\item \textbf{Within-batch sequence correlation.} Reports were split into non-overlapping 32-token windows; multiple windows from the same report may co-occur within a single training batch. This violates the i.i.d.\ assumption implicit in our bin-collision analysis and may either inflate or deflate empirical reconstruction success relative to the theoretical per-token collision rate. A correlation-aware extension of the analytic inversion bound is left to future work.

\item \textbf{Clinical-entity recall scope.} The MedGemma NER analysis was applied to a saturation-justified sample of 100 reconstructed discharge summaries (Dischargesum); extension of this analysis to the Diagnosis and MIMIC-CXR corpora --- which differ in report length, vocabulary density, and clinical specialty --- was beyond the compute budget of the present study and is left for future work. The clinical-entity leakage claims should therefore be interpreted as established for the discharge-summary regime and as a hypothesis to be tested in the other two report types.

\item \textbf{Re-execution for the paired t-test analysis.} The paired t-test analysis reported in Multimedia Appendix 8 is computed from an independent re-execution of the experimental grid. The primary reported reconstruction values in Appendices 1 and 2 are those from the original five-run experiment, on which all substantive conclusions of the manuscript rely. The re-execution reproduces the largest-batch ($B = 256$) sBLEU values within 0.05 of the originally reported Appendix 2 values in 8 of 9 (tokenizer, dataset) cells (the RadBERT / MIMIC-CXR cell differs by 0.064) and preserves the declining-with-batch-size trend; however, at smaller batches the re-execution yields exact-sentence-accuracy values that differ substantially from those in Appendix 1 (for example, 64.69\% versus 42.1\% for GPT-2 on Discharge at $B = 64$), and it therefore should not be interpreted as reproducing the exact-sentence-accuracy values reported in the manuscript. Multimedia Appendix 8 is provided solely as the basis for the exact paired t-test analysis.

\item \textbf{Retrospective Attack Focus.} This study focused solely on post hoc attack analysis. No privacy-preserving defenses (e.g., differential privacy, secure aggregation, anomaly detection) were implemented or evaluated. While this allows for a clean assessment of leakage potential, it leaves open questions about mitigation feasibility and deployment cost.

\item \textbf{Tokenizer Isolation vs.\ Full Model Effects.} We controlled for model architecture to isolate the effect of tokenization on inversion risk. While informative, real-world systems often use both tokenizer and model co-adaptation. Thus, privacy risks may be higher or lower depending on the full pipeline design.

\item \textbf{Data partitioning.} Our experiments use an approximately IID partition across six simulated clients. Real-world cross-institutional federated deployments are typically non-IID --- site-specific protocols, demographic skew, and specialty-specific report distributions can introduce heterogeneity that affects gradient sparsity and consequently inversion vulnerability. The direction of this effect is not a priori clear (non-IID gradients carry stronger per-client signal but also higher variance), and a systematic non-IID evaluation is left for future work.

\item \textbf{Sequence length.} The 32-token regime adopted here is shorter than typical clinical narrative passages. Longer sequences increase the input dimensionality $m \times L$ of each per-sample reconstruction, which raises the number of bins $K$ required for unambiguous disentanglement and thereby reduces single-step reconstruction success in practice. Our reported leakage rates are therefore an upper bound for the 32-token regime; longer-sequence regimes are expected to be harder to attack with a single fixed $K$, all else equal. Extending the analysis to longer-sequence training is left for future work.

\item \textbf{Client count.} Our experiments fix the simulated client count at six, reflecting a typical small-consortium clinical collaboration. The analytic gradient-inversion attack characterized in this study operates on a single client's gradient at a single round, and the per-client per-round recovery rate is independent of the number of other clients participating in the federation. Aggregation effects across clients (which arise specifically under cryptographic secure aggregation or weighted averaging schemes that obscure individual client updates) are absent from this evaluation by design and are an explicit target for future defense-oriented work.

\item \textbf{Probe placement.} The imprint module is inserted immediately before the positional embedding so that token-level representations are exposed prior to any token-mixing layer (positional addition, attention). Placement at later layers would conflate tokenizer-segmentation effects with the dynamics of attention and feed-forward mixing, and would not be informative for the central question of this study. A systematic placement ablation across pre-embedding, post-embedding, and post-attention positions --- holding the tokenizer fixed --- would isolate the contribution of placement itself and is identified here as a target for future work.
\end{itemize}

Taken together, these limitations do not diminish the central contribution highlighting a structural vulnerability in federated clinical NLP but emphasize the need for prospective, defense-integrated evaluations in more realistic medical AI pipelines.

\section*{Future Work}
Several extensions to this study are warranted. First, systematic ablation of the imprint module's placement (pre-embedding, post-embedding, post-attention) would empirically validate the design rationale used here. Second, evaluation of differential privacy as a defense --- including utility--privacy trade-off curves at multiple $\varepsilon$ values, per-example clipping schedules, and interaction with secure aggregation --- is a substantial study in its own right and is the explicit subject of our planned follow-up work. Third, extension of clinical-entity recall to additional report types (Diagnosis, MIMIC-CXR) and to non-IID federated partitions would strengthen the external validity of the leakage estimates reported here.

\section*{Conclusions}
Federated learning holds great promise for enabling collaborative, multi-site radiology AI without centralizing patient records. However, our gradient-inversion experiments demonstrate that current FL pipelines can leak substantial portions of patient text, raising significant concerns under HIPAA and GDPR standards. In worst-case scenarios, up to 44\% of sentences were exactly reconstructed from shared gradients, and 18\% of clinical terms, including diagnoses, procedures, and medications, were recovered from the reconstructed text.

Importantly, domain-adapted tokenizers such as RadBERT, while improving semantic fidelity, also showed greater vulnerability to data reconstruction attacks compared to general-purpose alternatives like GPT-2 and LLaMA-2. This highlights a critical trade-off between clinical performance and privacy risk.

To ensure safe deployment of federated clinical models, radiology departments and AI developers should rigorously evaluate privacy vulnerabilities in real-world settings, balancing utility with exposure risk. Regulatory bodies and standardization agencies may consider establishing clear audit pathways and certification frameworks tailored to the unique threats posed by clinical text.

\section*{Acknowledgements}
This work was supported by Grant Number 01ZZ2316A-O. The authors thank Maximilian Zenk (Division of Medical Image Computing, DKFZ) for his valuable insights during manuscript preparation. We also appreciate the guidance and prior work of Jonas Geiping and Liam Fowl in developing federated transformer methods. We acknowledge the developers of the breaching library for providing the gradient-inversion framework used in this study. Generative AI tools were used to assist with language editing, grammar checking, and formatting. All scientific content, analyses, and conclusions are the authors' own.

\section*{Funding Statement}
This work was partially supported by the PrivateAIM project, funded under the Medical Informatics Initiative by the German Federal Ministry of Education and Research (funding code 01ZZ2316A-O). The funder had no involvement in the study design, data collection, analysis, interpretation, or writing of the manuscript.

\section*{Conflicts of Interest}
The authors declare that they have no competing interests.

\section*{Data Availability}
The radiology report corpora analyzed in this study are publicly available: the Dischargesum dataset and the MIMIC-CXR via PhysioNet. The code implementing the federated learning experiments and gradient inversion attacks is available from the corresponding author upon reasonable request. All other data generated or analyzed during this study are included in this published article and its supplementary information files.

\section*{Authors' Contributions}
SP: Conceptualization, Methodology, Software, Validation, Formal Analysis, Investigation, Data Curation, Writing --- Original Draft, Writing --- Review \& Editing, Visualization, Project Administration. AM: Visualization, Writing --- Review \& Editing. DB: Writing --- Review \& Editing. SS: Resources, Supervision, Writing --- Review \& Editing. KMH: Supervision, Writing --- Review \& Editing. RF: Conceptualization, Resources, Supervision, Writing --- Review \& Editing.

\section*{Abbreviations}
\begin{itemize}[leftmargin=*,itemsep=0pt]
\item AI: artificial intelligence
\item BLEU: bilingual evaluation understudy
\item DP: differential privacy
\item FL: federated learning
\item G-BLEU: global BLEU
\item GDPR: General Data Protection Regulation
\item HIPAA: Health Insurance Portability and Accountability Act
\item LLM: large language model
\item PHI: protected health information
\item ROUGE-L: recall-oriented understudy for gisting evaluation
\end{itemize}

\section*{References}
\begin{enumerate}[label={[\arabic*]}, leftmargin=*, itemsep=1pt]
\item Castillo C, Steffens T, Sim L, Caffery L. The effect of clinical information on radiology reporting: A systematic review. J Med Radiat Sci Wiley; 2021 Mar;68(1):60--74.
\item Thirunavukarasu AJ, Ting DSJ, Elangovan K, Gutierrez L, Tan TF, Ting DSW. Large language models in medicine. Nat Med Nature Publishing Group US New York; 2023;29(8):1930--1940.
\item Radford A, Narasimhan K. Improving language understanding by generative pre-training. 2018. Available from: \url{https://api.semanticscholar.org/CorpusID:49313245}
\item Lecler A, Duron L, Soyer P. Revolutionizing radiology with GPT-based models: current applications, future possibilities and limitations of ChatGPT. Diagn Interv Imaging Elsevier; 2023;104(6):269--274.
\item Hu D, Zhang S, Liu Q, Zhu X, Liu B. Large language models in summarizing radiology report impressions for lung cancer in Chinese: evaluation study. J Med Internet Res 2025 Apr 3;27:e65547. doi: 10.2196/65547
\item Wu Q, Wu Q, Li H, Wang Y, Bai Y, Wu Y, Yu X, Li X, Dong P, Xue J, Shen D, Wang M. Evaluating large language models for automated reporting and data systems categorization: cross-sectional study. JMIR Med Inform 2024 July 17;12:e55799. doi: 10.2196/55799
\item Shool S, Adimi S, Saboori Amleshi R, Bitaraf E, Golpira R, Tara M. A systematic review of large language model (LLM) evaluations in clinical medicine. BMC Med Inform Decis Mak 2025 Mar 7;25(1):117. doi: 10.1186/s12911-025-02954-4
\item hhs.gov. The HIPAA Privacy Rule. Available from: \url{https://www.hhs.gov/hipaa/for-professionals/privacy/index.html}
\item gdpr.eu. General Data Protection Regulation (GDPR). Available from: \url{https://gdpr.eu/tag/gdpr/}
\item McMahan B, Moore E, Ramage D, Hampson S, y Arcas BA. Communication-efficient learning of deep networks from decentralized data. Artif Intell Stat PMLR; 2017. p. 1273--1282.
\item Sun Y, Khor HG, Wang Y, Wang Z, Zhao H, Zhang Y, Ma L, Zheng Z, Liao H. Continually tuning a large language model for multi-domain radiology report generation. Int Conf Med Image Comput Comput-Assist Interv Springer; 2024. p. 177--187.
\item Bhayana R. Chatbots and large language models in radiology: a practical primer for clinical and research applications. Radiology Radiological Society of North America; 2024;310(1):e232756.
\item D'Antonoli TA, Stanzione A, Bluethgen C, Vernuccio F, Ugga L, Klontzas ME, Cuocolo R, Cannella R, Ko\c{c}ak B. Large language models in radiology: fundamentals, applications, ethical considerations, risks, and future directions. Diagn Interv Radiol 2024;30(2):80.
\item Wong IN, Monteiro O, Baptista-Hon DT, Wang K, Lu W, Sun Z, Nie S, Yin Y. Leveraging foundation and large language models in medical artificial intelligence. Chin Med J (Engl) 2024;137(21):2529--2539.
\item Parampottupadam S, Floca R, Bounias D, Hamm B, Roy S, Sav S, Zenk M, Maier-Hein K. Client security alone fails in federated learning: 2D and 3D attack insights. MICCAI EMERGE Workshop 2024, Marrakesh, Morocco. Springer; 2024. p. 235--244.
\item Brauneck A, Schmalhorst L, Kazemi Majdabadi MM, Bakhtiari M, V\"olker U, Baumbach J, Baumbach L, Buchholtz G. Federated machine learning, privacy-enhancing technologies, and data protection laws in medical research: scoping review. J Med Internet Res 2023 Mar 30;25:e41588. doi: 10.2196/41588
\item Zhou J, Zhou L, Wang D, Xu X, Li H, Chu Y, Han W, Gao X. Personalized and privacy-preserving federated heterogeneous medical image analysis with PPPML-HMI. Comput Biol Med 2024 Feb;169:107861. doi: 10.1016/j.compbiomed.2023.107861
\item Kaissis G, Ziller A, Passerat-Palmbach J, Ryffel T, Usynin D, Trask A, Lima Jr I, Mancuso J, Jungmann F, Steinborn M-M, others. End-to-end privacy preserving deep learning on multi-institutional medical imaging. Nat Mach Intell Nature Publishing Group UK London; 2021;3(6):473--484.
\item Hatamizadeh A, Yin H, Molchanov P, Myronenko A, Li W, Dogra P, Feng A, Flores MG, Kautz J, Xu D, others. Do gradient inversion attacks make federated learning unsafe? IEEE Trans Med Imaging IEEE; 2023.
\item Akinci D'Antonoli T, Bluethgen C. A new era of text mining in radiology with privacy-preserving LLMs. Radiol Artif Intell Radiological Society of North America; 2024;6(4):e240261.
\item Gu Y, Tinn R, Cheng H, Lucas M, Usuyama N, Liu X, Naumann T, Gao J, Poon H. Domain-specific language model pretraining for biomedical natural language processing. ACM Trans Comput Healthc 2021;3(1):1--23.
\item Zhang X, Tian C, Yang X, Chen L, Li Z, Petzold LR. Alpacare: Instruction-tuned large language models for medical application. arXiv preprint arXiv:2310.14558; 2023. Available from: \url{https://arxiv.org/abs/2310.14558}
\item Yan A, McAuley J, Lu X, Du J, Chang EY, Gentili A, Hsu C-N. RadBERT: adapting transformer-based language models to radiology. Radiol Artif Intell 2022 July 1;4(4):e210258. doi: 10.1148/ryai.210258
\item Solaiman I, Brundage M, Clark J, Askell A, Herbert-Voss A, Wu J, Radford A, Wang J. Release strategies and the social impacts of language models. arXiv preprint arXiv:1908.09203 [cs.CL]; 2019. Available from: \url{https://arxiv.org/abs/1908.09203}
\item Touvron H, Martin L, Stone K, Albert P, Almahairi A, Babaei Y, Bashlykov N, Batra S, Bhargava P, Bhosale S, others. Llama 2: Open foundation and fine-tuned chat models. arXiv preprint arXiv:2307.09288; 2023. Available from: \url{https://arxiv.org/abs/2307.09288}
\item Zhao B, Mopuri KR, Bilen H. iDLG: improved deep leakage from gradients. arXiv; 2020. doi: 10.48550/arXiv.2001.02610
\item Sellergren A, Kazemzadeh S, Jaroensri T, Kiraly A, Traverse M, Kohlberger T, Xu S, Jamil F, Hughes C, Lau C, others. MedGemma technical report. arXiv; 2025. doi: 10.48550/arXiv.2507.05201
\item Peng C. Dischargesum Dataset. 2024. Available from: \url{https://huggingface.co/dischargesum}
\item Johnson AEW, Pollard TJ, Greenbaum NR, Lungren MP, Deng C, Peng Y, Lu Z, Mark RG, Berkowitz SJ, Horng S. MIMIC-CXR-JPG, a large publicly available database of labeled chest radiographs. 2019. Available from: \url{https://arxiv.org/abs/1901.07042}
\item Papineni K, Roukos S, Ward T, Zhu W-J. BLEU: a method for automatic evaluation of machine translation. Proc 40th Annu Meet Assoc Comput Linguist USA: Association for Computational Linguistics; 2002. p. 311--318. doi: 10.3115/1073083.1073135
\item Delbrouck J-B, Varma M, Chambon P, Langlotz C. Overview of the RadSum23 shared task on multi-modal and multi-anatomical radiology report summarization. 22nd Workshop Biomed Nat Lang Process BioNLP Shar Tasks Toronto, Canada: Association for Computational Linguistics; 2023. p. 478--482. doi: 10.18653/v1/2023.bionlp-1.45
\item Veen DV, Uden CV, Blankemeier L, Delbrouck J-B, Aali A, Bluethgen C, Pareek A, Polacin M, Reis EP, Seehofnerova A, Rohatgi N, Hosamani P, Collins W, Ahuja N, Langlotz C, Hom J, Gatidis S, Pauly J, Chaudhari A. Clinical text summarization: adapting large language models can outperform human experts. In Review; 2023. doi: 10.21203/rs.3.rs-3483777/v1
\item Lin C-Y. ROUGE: a package for automatic evaluation of summaries. Annu Meet Assoc Comput Linguist 2004. Available from: \url{https://api.semanticscholar.org/CorpusID:964287}
\item Dwork C. Differential Privacy. In: Bugliesi M, Preneel B, Sassone V, Wegener I, editors. Autom Lang Program Berlin, Heidelberg: Springer Berlin Heidelberg; 2006. p. 1--12.
\item Bonawitz K, Ivanov V, Kreuter B, Marcedone A, McMahan HB, Patel S, Ramage D, Segal A, Seth K. Practical secure aggregation for privacy-preserving machine learning. Proc 2017 ACM SIGSAC Conf Comput Commun Secur New York, NY, USA: Association for Computing Machinery; 2017. p. 1175--1191. doi: 10.1145/3133956.3133982
\item Zhang X, Kang Y, Chen K, Fan L, Yang Q. Trading off privacy, utility, and efficiency in federated learning. ACM Trans Intell Syst Technol 2023 Dec 31;14(6):1--32. doi: 10.1145/3595185
\item Akkus A, Aghdam MP, Li M, Chu J, Backes M, Zhang Y, Sav S. Generated data with fake privacy: hidden dangers of fine-tuning large language models on generated data. arXiv; 2024. doi: 10.48550/arXiv.2409.11423
\item Stadler T, Oprisanu B, Troncoso C. Synthetic data --- anonymisation groundhog day. arXiv; 2020. doi: 10.48550/arXiv.2011.07018
\item Jayaraman B, Evans D. Evaluating differentially private machine learning in practice. arXiv; 2019. doi: 10.48550/arXiv.1902.08874
\item Hosseini SM, Sikaroudi M, Babaei M, Tizhoosh HR. Cluster based secure multi-party computation in federated learning for histopathology images. arXiv; 2022. doi: 10.48550/arXiv.2208.10919
\item Kaabachi B, Despraz J, Meurers T, Otte K, Halilovic M, Kulynych B, Prasser F, Raisaro JL. A scoping review of privacy and utility metrics in medical synthetic data. Npj Digit Med 2025 Jan 27;8(1):60. doi: 10.1038/s41746-024-01359-3
\item Fowl L, Geiping J, Czaja W, Goldblum M, Goldstein T. Robbing the Fed: directly obtaining private data in federated learning with modified models. International Conference on Learning Representations (ICLR) 2022; 2022. Available from: \url{https://arxiv.org/abs/2110.13057}
\item Geiping J, Bauermeister H, Dr\"oge H, Moeller M. Inverting gradients --- how easy is it to break privacy in federated learning? Advances in Neural Information Processing Systems (NeurIPS) 2020; 2020. Available from: \url{https://arxiv.org/abs/2003.14053}
\item Kairouz P, McMahan HB, Avent B, et al. Advances and open problems in federated learning. Foundations and Trends in Machine Learning, vol.\ 14, no.\ 1-2; 2021. doi: 10.1561/2200000083
\item Boenisch F, Dziedzic A, Schuster R, Shamsabadi AS, Shumailov I, Papernot N. When the curious abandon honesty: federated learning is not private. IEEE European Symposium on Security and Privacy (EuroS\&P) 2023; 2023.
\item Deng J, Wang Y, Li J, Shang C, Qin C, Liu H, Rajasekaran S, Ding C. TAG: gradient attack on transformer-based language models. Findings of EMNLP 2021; 2021. doi: 10.48550/arXiv.2103.06819
\item Zhu L, Liu Z, Han S. Deep leakage from gradients. Advances in Neural Information Processing Systems (NeurIPS) 2019; 2019. doi: 10.48550/arXiv.1906.08935
\end{enumerate}

\clearpage
\appendix
\section*{Multimedia Appendices}
\addcontentsline{toc}{section}{Multimedia Appendices}

\subsection*{Multimedia Appendix 1}
\textbf{Exact sentence-level reconstruction accuracy (\%) across batch sizes (64, 128, 256) for three tokenizers (GPT-2, RadBERT, and LLaMA-2) on three radiology datasets (Discharge, Diagnosis, and MIMIC-CXR).} Accuracy is computed as the proportion of sentences perfectly reconstructed from shared gradients.

\begin{table}[H]
\centering
\begin{tabular}{@{}llrrr@{}}
\toprule
\textbf{Tokenizer (vocab size)} & \textbf{Dataset} & \textbf{Batch 64} & \textbf{Batch 128} & \textbf{Batch 256} \\
\midrule
\multirow{3}{*}{GPT-2 (50,257)} & Discharge & 42.1 & 40.7 & 37.3 \\
 & Diagnosis & 41.4 & 38.0 & 34.6 \\
 & MIMIC-CXR & 40.5 & 36.5 & 32.5 \\
\midrule
\multirow{3}{*}{RadBERT (30,522)} & Discharge & 42.3 & 40.0 & 37.2 \\
 & Diagnosis & 41.5 & 39.0 & 35.1 \\
 & MIMIC-CXR & 43.5 & 40.6 & 36.3 \\
\midrule
\multirow{3}{*}{LLaMA-2 (32,000)} & Discharge & 39.4 & 37.0 & 34.3 \\
 & Diagnosis & 38.5 & 35.4 & 31.4 \\
 & MIMIC-CXR & 39.0 & 34.7 & 30.6 \\
\bottomrule
\end{tabular}
\end{table}

\subsection*{Multimedia Appendix 2}
\textbf{Mean reconstruction scores for S-BLEU (S), G-BLEU (G), and ROUGE-L (R) across batch sizes (64, 128, 256) for three tokenizers on three radiology datasets.} Higher scores indicate greater recovery of clinical phrasing.

\begin{table}[H]
\centering
\begin{tabular}{@{}llccc@{}}
\toprule
\textbf{Dataset} & \textbf{Tokenizer} & \textbf{64 (S/G/R)} & \textbf{128 (S/G/R)} & \textbf{256 (S/G/R)} \\
\midrule
\multirow{3}{*}{Discharge} & GPT-2 & 0.44 / 0.41 / 0.41 & 0.38 / 0.37 / 0.38 & 0.33 / 0.33 / 0.32 \\
 & RadBERT & 0.48 / 0.46 / 0.45 & 0.42 / 0.40 / 0.39 & 0.35 / 0.34 / 0.33 \\
 & LLaMA-2 & 0.39 / 0.36 / 0.39 & 0.35 / 0.33 / 0.36 & 0.30 / 0.29 / 0.33 \\
\midrule
\multirow{3}{*}{Diagnosis} & GPT-2 & 0.43 / 0.40 / 0.40 & 0.37 / 0.36 / 0.37 & 0.32 / 0.31 / 0.30 \\
 & RadBERT & 0.47 / 0.45 / 0.44 & 0.41 / 0.39 / 0.38 & 0.34 / 0.33 / 0.32 \\
 & LLaMA-2 & 0.38 / 0.35 / 0.38 & 0.34 / 0.32 / 0.34 & 0.29 / 0.28 / 0.31 \\
\midrule
\multirow{3}{*}{MIMIC-CXR} & GPT-2 & 0.44 / 0.39 / 0.39 & 0.36 / 0.35 / 0.36 & 0.31 / 0.30 / 0.29 \\
 & RadBERT & 0.52 / 0.44 / 0.43 & 0.46 / 0.38 / 0.37 & 0.38 / 0.32 / 0.31 \\
 & LLaMA-2 & 0.40 / 0.34 / 0.37 & 0.33 / 0.31 / 0.33 & 0.28 / 0.27 / 0.30 \\
\bottomrule
\end{tabular}
\end{table}

\subsection*{Multimedia Appendix 3}
\textbf{Distribution of exact sentence-level reconstruction accuracy across datasets, tokenizers, and batch sizes} (violin plots over the five independent runs).

\begin{figure}[H]
\centering
\includegraphics[width=\linewidth]{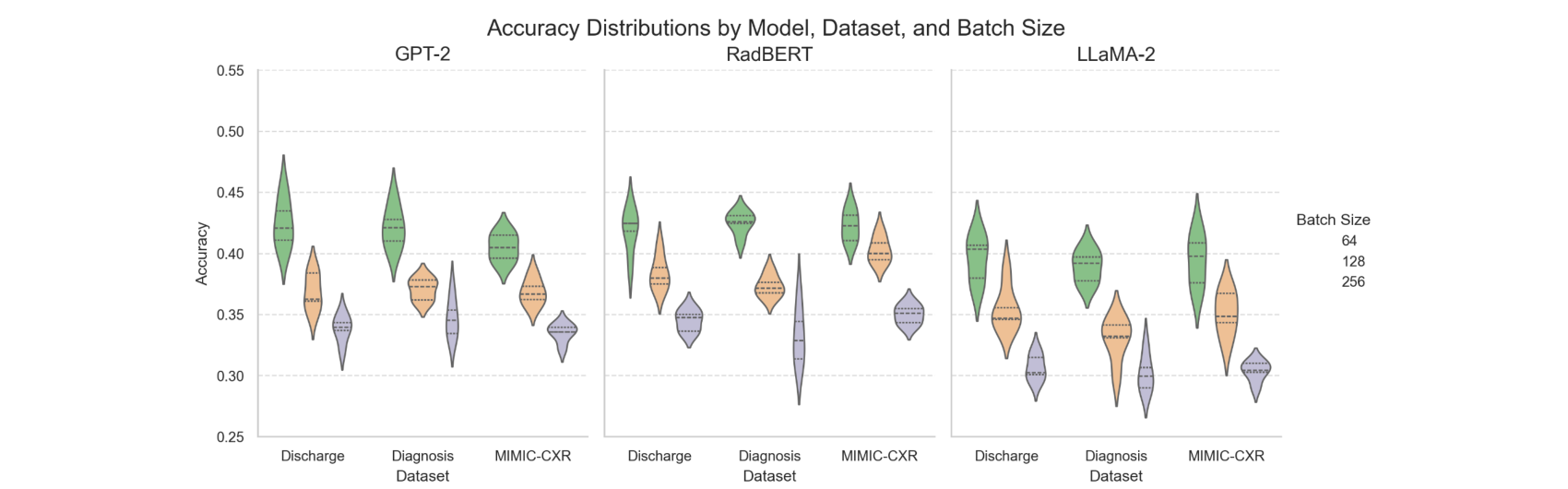}
\end{figure}

\subsection*{Multimedia Appendix 4}
\textbf{Distribution of ROUGE-L scores across datasets, tokenizers, and batch sizes} (violin plots over the five independent runs).

\begin{figure}[H]
\centering
\includegraphics[width=\linewidth]{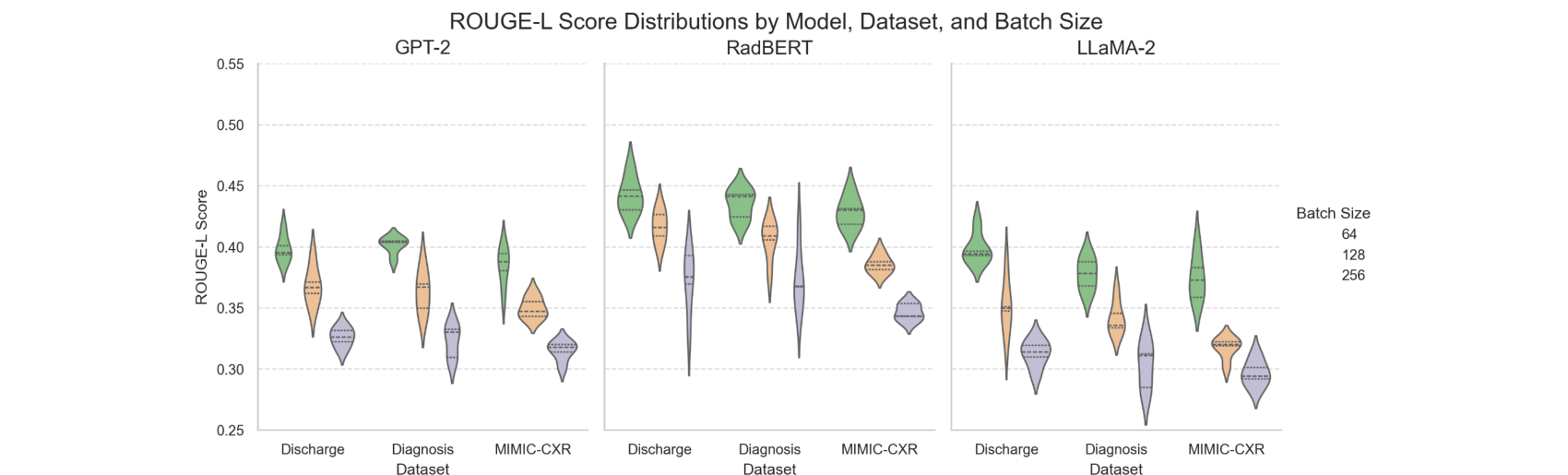}
\end{figure}

\subsection*{Multimedia Appendix 5}
\textbf{Distribution of G-BLEU scores across datasets, tokenizers, and batch sizes} (violin plots over the five independent runs).

\begin{figure}[H]
\centering
\includegraphics[width=0.62\linewidth]{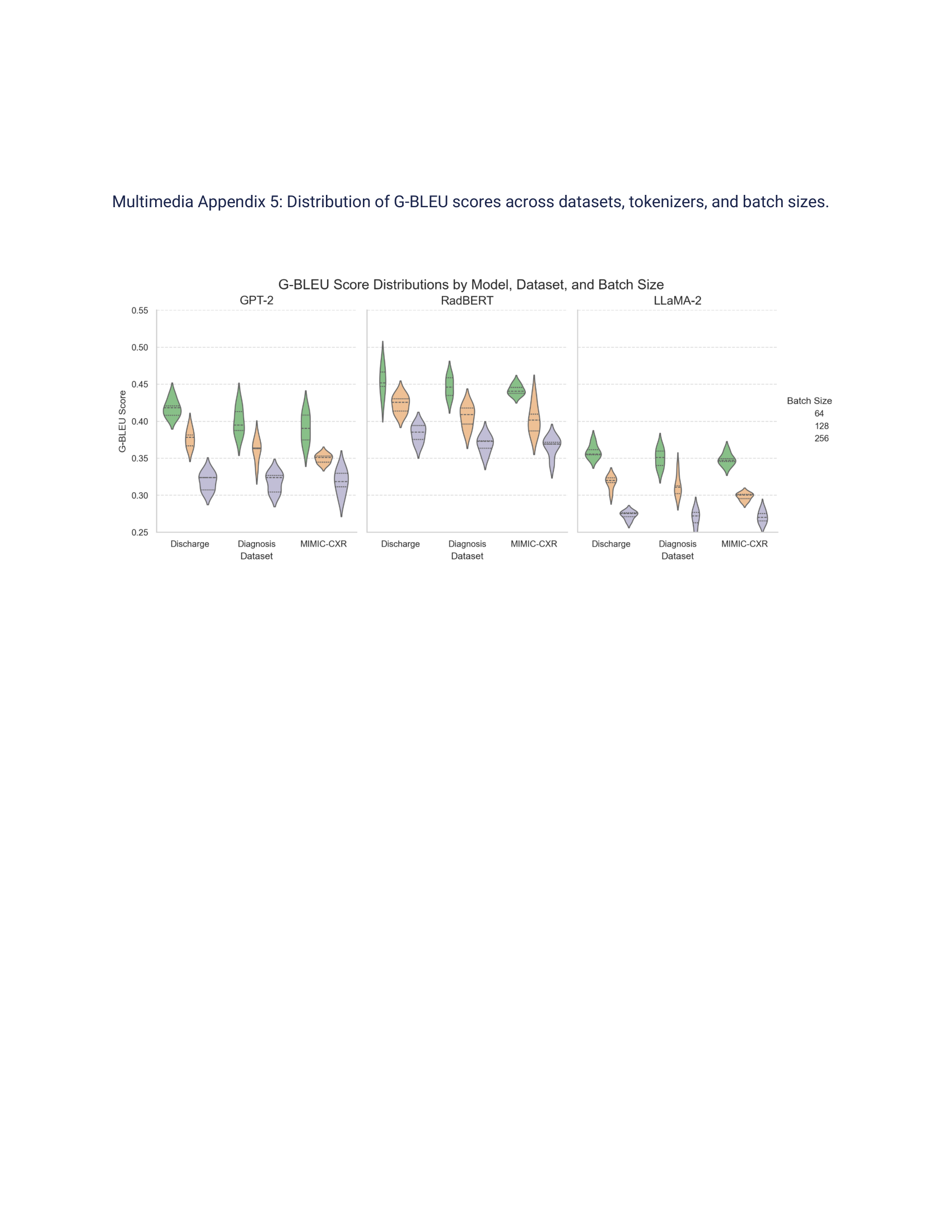}
\end{figure}

\clearpage
\subsection*{Multimedia Appendix 6}
\textbf{95\% confidence intervals for exact sentence-level reconstruction accuracy across batch sizes and tokenizers,} derived from the five-run experiment underlying the main Results.

Confidence intervals were computed using Student's $t$-distribution with $n = 5$ independent runs per cell (degrees of freedom = 4, critical value $t_{0.025} = 2.776$). The standard deviation upper bound of 1 percentage point was used, as reported in the main Results (Section \textit{Quantitative Reconstruction Success}); where the true per-cell standard deviation is below this upper bound, reported CI half-widths are conservative. Means are reproduced exactly from Multimedia Appendix 1.

\begin{table}[H]
\centering
\begin{tabular}{@{}lllrl@{}}
\toprule
\textbf{Tokenizer} & \textbf{Dataset} & \textbf{Batch size} & \textbf{Mean (\%)} & \textbf{95\% CI} \\
\midrule
\multirow{9}{*}{GPT-2} & \multirow{3}{*}{Discharge} & 64 & 42.10 & [40.86, 43.34] \\
 & & 128 & 40.70 & [39.46, 41.94] \\
 & & 256 & 37.30 & [36.06, 38.54] \\
\cmidrule(l){2-5}
 & \multirow{3}{*}{Diagnosis} & 64 & 41.40 & [40.16, 42.64] \\
 & & 128 & 38.00 & [36.76, 39.24] \\
 & & 256 & 34.60 & [33.36, 35.84] \\
\cmidrule(l){2-5}
 & \multirow{3}{*}{MIMIC-CXR} & 64 & 40.50 & [39.26, 41.74] \\
 & & 128 & 36.50 & [35.26, 37.74] \\
 & & 256 & 32.50 & [31.26, 33.74] \\
\midrule
\multirow{9}{*}{RadBERT} & \multirow{3}{*}{Discharge} & 64 & 42.30 & [41.06, 43.54] \\
 & & 128 & 40.00 & [38.76, 41.24] \\
 & & 256 & 37.20 & [35.96, 38.44] \\
\cmidrule(l){2-5}
 & \multirow{3}{*}{Diagnosis} & 64 & 41.50 & [40.26, 42.74] \\
 & & 128 & 39.00 & [37.76, 40.24] \\
 & & 256 & 35.10 & [33.86, 36.34] \\
\cmidrule(l){2-5}
 & \multirow{3}{*}{MIMIC-CXR} & 64 & 43.50 & [42.26, 44.74] \\
 & & 128 & 40.60 & [39.36, 41.84] \\
 & & 256 & 36.30 & [35.06, 37.54] \\
\midrule
\multirow{9}{*}{LLaMA-2} & \multirow{3}{*}{Discharge} & 64 & 39.40 & [38.16, 40.64] \\
 & & 128 & 37.00 & [35.76, 38.24] \\
 & & 256 & 34.30 & [33.06, 35.54] \\
\cmidrule(l){2-5}
 & \multirow{3}{*}{Diagnosis} & 64 & 38.50 & [37.26, 39.74] \\
 & & 128 & 35.40 & [34.16, 36.64] \\
 & & 256 & 31.40 & [30.16, 32.64] \\
\cmidrule(l){2-5}
 & \multirow{3}{*}{MIMIC-CXR} & 64 & 39.00 & [37.76, 40.24] \\
 & & 128 & 34.70 & [33.46, 35.94] \\
 & & 256 & 30.60 & [29.36, 31.84] \\
\bottomrule
\end{tabular}
\end{table}

\subsection*{Multimedia Appendix 7}
\textbf{Training details and reproducibility configuration.} Hyperparameters and infrastructure-relevant settings used for all experiments reported in this study.

\begin{table}[H]
\centering
\begin{tabular}{@{}>{\RaggedRight}p{5.6cm}>{\RaggedRight\arraybackslash}p{9cm}@{}}
\toprule
\textbf{Parameter} & \textbf{Value} \\
\midrule
Optimizer & AdamW \\
Learning rate & $5 \times 10^{-4}$ \\
Local batch size (sequences) & 64, 128, 256 \\
Sequence length & 32 tokens \\
FL rounds simulated per measurement & 1 (single-step attack per gradient inversion measurement) \\
Aggregation rule & FedAvg (logical; per-client per-round inversion) \\
Number of simulated clients & 6 (uniform random partition; IID) \\
Independent runs per configuration & 5 (different random seeds) \\
Foundation model & 3-layer Transformer encoder; $m = 96$, 8 heads, FFN 1536; $\approx$ 13.4\,M parameters \\
Tokenizers & GPT-2 (50,257 tokens); StanfordAIMI RadBERT (30,522 tokens); LLaMA-2-7B (32,000 tokens) \\
Imprint bins ($K$) & 1,000 \\
Imprint linear-function init & Random Gaussian (\texttt{linfunc = randn}), gain 1.0 \\
Imprint bias init & Cumulative Laplace quantile thresholds \\
Numerical stability ridge $\varepsilon$ & $10^{-6}$ \\
Hardware & NVIDIA A100 (40 GB), single-GPU per run \\
Software & Python 3.10, PyTorch 2.x, HuggingFace Transformers, breaching library [A1] \\
Random seeds & $\{0, 1, 2, 3, 4\}$ \\
\bottomrule
\end{tabular}
\end{table}

\noindent\textit{Appendix reference.}\\
{[}A1{]} Geiping J, Fowl L, Huang WR, Czaja W, Taylor G, Moeller M, Goldstein T. Decepticons: corrupted transformers breach privacy in federated learning for language models. arXiv preprint arXiv:2201.12675; 2022. Available from: \url{https://arxiv.org/abs/2201.12675}

\clearpage
\subsection*{Multimedia Appendix 8}
\textbf{Verification Re-run --- Cell Means and Exact Paired t-tests.}

This appendix reports a re-execution of the experimental grid. The re-execution covers the full grid (3 tokenizers $\times$ 3 batch sizes $\times$ 3 datasets $\times$ 5 seeds = 135 attack runs) and uses the same attack pipeline as the originally reported experiment (the imprint module of the breaching library, with $K = 1000$ bins, 32-token sequences, and per-position cumulative-bin disentanglement). The per-run output is retained in a format suitable for direct paired-$t$-test computation. Seeds 0--4 were used for the re-execution. Cell means are reported in the first table below; exact paired $t$-tests on the per-seed exact-sentence reconstruction accuracy (in percentage points) are reported in the second table.

\begin{table}[H]
\footnotesize
\centering
\setlength{\tabcolsep}{4pt}
\caption*{\textbf{Table A8-1.} Cell means ($n = 5$ seeds) for exact sentence-level reconstruction accuracy, sBLEU, and ROUGE-L, by (tokenizer, batch size, dataset). All cells $n = 5$ seeds.}
\begin{tabular}{@{}llrrrrrrr@{}}
\toprule
 & & & \multicolumn{2}{c}{\textbf{Exact (\%)}} & \multicolumn{2}{c}{\textbf{sBLEU}} & \multicolumn{2}{c}{\textbf{ROUGE-L}} \\
\cmidrule(lr){4-5}\cmidrule(lr){6-7}\cmidrule(l){8-9}
\textbf{Tokenizer} & \textbf{Batch} & \textbf{Dataset} & \textbf{Mean} & \textbf{SD} & \textbf{Mean} & \textbf{SD} & \textbf{Mean} & \textbf{SD} \\
\midrule
GPT-2 & 64 & Discharge & 64.69 & 3.60 & 0.662 & 0.026 & 0.667 & 0.026 \\
GPT-2 & 64 & MIMIC-CXR & 74.69 & 6.67 & 0.752 & 0.068 & 0.755 & 0.068 \\
GPT-2 & 64 & Diagnosis & 69.06 & 4.04 & 0.700 & 0.041 & 0.705 & 0.039 \\
GPT-2 & 128 & Discharge & 46.41 & 6.50 & 0.486 & 0.055 & 0.497 & 0.054 \\
GPT-2 & 128 & MIMIC-CXR & 49.84 & 2.73 & 0.518 & 0.025 & 0.526 & 0.022 \\
GPT-2 & 128 & Diagnosis & 47.03 & 1.50 & 0.495 & 0.018 & 0.505 & 0.019 \\
GPT-2 & 256 & Discharge & 27.34 & 1.66 & 0.311 & 0.022 & 0.329 & 0.021 \\
GPT-2 & 256 & MIMIC-CXR & 28.67 & 2.60 & 0.316 & 0.024 & 0.328 & 0.021 \\
GPT-2 & 256 & Diagnosis & 27.19 & 2.00 & 0.310 & 0.016 & 0.325 & 0.013 \\
\midrule
LLaMA-2 & 64 & Discharge & 67.50 & 4.61 & 0.686 & 0.045 & 0.690 & 0.047 \\
LLaMA-2 & 64 & MIMIC-CXR & 68.44 & 7.02 & 0.696 & 0.068 & 0.699 & 0.069 \\
LLaMA-2 & 64 & Diagnosis & 65.94 & 5.57 & 0.666 & 0.050 & 0.670 & 0.047 \\
LLaMA-2 & 128 & Discharge & 48.44 & 2.34 & 0.508 & 0.021 & 0.516 & 0.023 \\
LLaMA-2 & 128 & MIMIC-CXR & 49.53 & 4.67 & 0.511 & 0.044 & 0.519 & 0.042 \\
LLaMA-2 & 128 & Diagnosis & 50.16 & 3.76 & 0.520 & 0.032 & 0.530 & 0.031 \\
LLaMA-2 & 256 & Discharge & 27.50 & 1.50 & 0.303 & 0.012 & 0.316 & 0.010 \\
LLaMA-2 & 256 & MIMIC-CXR & 29.53 & 2.32 & 0.320 & 0.019 & 0.332 & 0.016 \\
LLaMA-2 & 256 & Diagnosis & 27.58 & 3.66 & 0.309 & 0.026 & 0.325 & 0.020 \\
\midrule
RadBERT & 64 & Discharge & 70.00 & 6.85 & 0.711 & 0.053 & 0.716 & 0.050 \\
RadBERT & 64 & MIMIC-CXR & 67.50 & 3.39 & 0.680 & 0.030 & 0.685 & 0.028 \\
RadBERT & 64 & Diagnosis & 67.50 & 5.34 & 0.679 & 0.054 & 0.682 & 0.053 \\
RadBERT & 128 & Discharge & 50.78 & 1.83 & 0.532 & 0.020 & 0.537 & 0.024 \\
RadBERT & 128 & MIMIC-CXR & 48.91 & 2.80 & 0.503 & 0.022 & 0.511 & 0.022 \\
RadBERT & 128 & Diagnosis & 46.41 & 3.25 & 0.482 & 0.020 & 0.492 & 0.017 \\
RadBERT & 256 & Discharge & 28.52 & 1.77 & 0.319 & 0.018 & 0.336 & 0.017 \\
RadBERT & 256 & MIMIC-CXR & 28.83 & 2.66 & 0.316 & 0.020 & 0.329 & 0.017 \\
RadBERT & 256 & Diagnosis & 26.64 & 2.49 & 0.302 & 0.022 & 0.317 & 0.022 \\
\bottomrule
\end{tabular}
\end{table}

\begin{table}[H]
\footnotesize
\centering
\caption*{\textbf{Table A8-2.} Exact paired $t$-tests on per-seed exact-sentence reconstruction accuracy (percentage points), by (batch size, dataset, tokenizer pair). Significance key: *** $p < 0.001$, ** $p < 0.01$, * $p < 0.05$, ns $p \ge 0.05$.}
\begin{tabular}{@{}llrrrrrl@{}}
\toprule
\textbf{Batch} & \textbf{Dataset} & \textbf{Pair} & \textbf{$n$} & \textbf{$\Delta$ (pp)} & \textbf{SD diff (pp)} & \textbf{$t$} & \textbf{$p$} \\
\midrule
64 & Discharge & GPT-2 vs LLaMA-2 & 5 & $-2.81$ & 5.46 & $-1.15$ & .3134 (ns) \\
64 & Discharge & GPT-2 vs RadBERT & 5 & $-5.31$ & 7.78 & $-1.53$ & .2016 (ns) \\
64 & Discharge & LLaMA-2 vs RadBERT & 5 & $-2.50$ & 10.63 & $-0.53$ & .6268 (ns) \\
64 & MIMIC-CXR & GPT-2 vs LLaMA-2 & 5 & $+6.25$ & 7.89 & $+1.77$ & .1512 (ns) \\
64 & MIMIC-CXR & GPT-2 vs RadBERT & 5 & $+7.19$ & 5.70 & $+2.82$ & \textbf{.0478 (*)} \\
64 & MIMIC-CXR & LLaMA-2 vs RadBERT & 5 & $+0.94$ & 7.94 & $+0.26$ & .8047 (ns) \\
64 & Diagnosis & GPT-2 vs LLaMA-2 & 5 & $+3.12$ & 6.35 & $+1.10$ & .3327 (ns) \\
64 & Diagnosis & GPT-2 vs RadBERT & 5 & $+1.56$ & 7.97 & $+0.44$ & .6836 (ns) \\
64 & Diagnosis & LLaMA-2 vs RadBERT & 5 & $-1.56$ & 7.57 & $-0.46$ & .6686 (ns) \\
\midrule
128 & Discharge & GPT-2 vs LLaMA-2 & 5 & $-2.03$ & 6.04 & $-0.75$ & .4940 (ns) \\
128 & Discharge & GPT-2 vs RadBERT & 5 & $-4.38$ & 7.94 & $-1.23$ & .2854 (ns) \\
128 & Discharge & LLaMA-2 vs RadBERT & 5 & $-2.34$ & 2.92 & $-1.79$ & .1475 (ns) \\
128 & MIMIC-CXR & GPT-2 vs LLaMA-2 & 5 & $+0.31$ & 7.26 & $+0.10$ & .9279 (ns) \\
128 & MIMIC-CXR & GPT-2 vs RadBERT & 5 & $+0.94$ & 2.30 & $+0.91$ & .4144 (ns) \\
128 & MIMIC-CXR & LLaMA-2 vs RadBERT & 5 & $+0.62$ & 6.14 & $+0.23$ & .8310 (ns) \\
128 & Diagnosis & GPT-2 vs LLaMA-2 & 5 & $-3.12$ & 3.27 & $-2.14$ & .0993 (ns) \\
128 & Diagnosis & GPT-2 vs RadBERT & 5 & $+0.62$ & 3.60 & $+0.39$ & .7174 (ns) \\
128 & Diagnosis & LLaMA-2 vs RadBERT & 5 & $+3.75$ & 6.33 & $+1.32$ & .2560 (ns) \\
\midrule
256 & Discharge & GPT-2 vs LLaMA-2 & 5 & $-0.16$ & 2.50 & $-0.14$ & .8954 (ns) \\
256 & Discharge & GPT-2 vs RadBERT & 5 & $-1.17$ & 1.07 & $-2.45$ & .0705 (ns) \\
256 & Discharge & LLaMA-2 vs RadBERT & 5 & $-1.02$ & 2.24 & $-1.02$ & .3675 (ns) \\
256 & MIMIC-CXR & GPT-2 vs LLaMA-2 & 5 & $-0.86$ & 4.57 & $-0.42$ & .6956 (ns) \\
256 & MIMIC-CXR & GPT-2 vs RadBERT & 5 & $-0.16$ & 3.53 & $-0.10$ & .9260 (ns) \\
256 & MIMIC-CXR & LLaMA-2 vs RadBERT & 5 & $+0.70$ & 3.40 & $+0.46$ & .6677 (ns) \\
256 & Diagnosis & GPT-2 vs LLaMA-2 & 5 & $-0.39$ & 2.99 & $-0.29$ & .7846 (ns) \\
256 & Diagnosis & GPT-2 vs RadBERT & 5 & $+0.55$ & 2.77 & $+0.44$ & .6817 (ns) \\
256 & Diagnosis & LLaMA-2 vs RadBERT & 5 & $+0.94$ & 5.21 & $+0.40$ & .7079 (ns) \\
\bottomrule
\end{tabular}
\end{table}

\noindent\textit{Comparison with originally reported values.} The largest-batch values ($B = 256$) reproduce within 0.05 sBLEU of the originally reported Appendix 2 values in 8 of 9 (tokenizer, dataset) cells (the RadBERT / MIMIC-CXR cell differs by 0.064), and the declining-with-batch-size trend is preserved throughout. At smaller batches ($B = 64$, 128) the re-execution produced sBLEU values somewhat higher than originally reported; we view this appendix as consistent with the qualitative findings of the manuscript rather than as a replacement for the originally reported reconstruction values, which remain in Appendices 1--2. Of 27 pairwise tokenizer comparisons at $\alpha = 0.05$, one is statistically significant. With $n = 5$ seeds per cell and observed per-cell standard deviation in the 2--7 percentage-point range, the power to detect tokenizer-level effect-size differences in the 2--4 pp range is limited; the absence of significance therefore reflects sampling power rather than absence of effect. Cell-level effect sizes preserve the manuscript's qualitative ordering at the largest batch on the Discharge dataset (RadBERT 28.52\%, GPT-2 27.34\%, LLaMA-2 27.50\% mean exact-sentence accuracy).

\end{document}